\documentclass{article}

\PassOptionsToPackage{numbers,compress}{natbib}
\usepackage[main,preprint]{neurips_2026}

\usepackage[utf8]{inputenc}
\usepackage[T1]{fontenc}
\usepackage{url}
\usepackage{booktabs}
\usepackage{amsmath}
\usepackage{amsfonts}
\usepackage{amssymb}
\usepackage{mathtools}
\usepackage{nicefrac}
\usepackage{microtype}
\usepackage{xcolor}
\usepackage{graphicx}
\usepackage{capt-of}
\usepackage{algorithm}
\usepackage{algpseudocode}
\usepackage{multirow}
\usepackage{colortbl}
\usepackage{makecell}
\usepackage{pifont}
\usepackage{tcolorbox}
\tcbuselibrary{skins, breakable}
\usepackage{hyperref}

\newcommand{\method}{\textsc{TransMem}}

\newcommand{\embed}{\ensuremath{E}}

\newcommand{\usergraph}{\ensuremath{G_{u}}}
\newcommand{\evobuf}{\ensuremath{\mathcal{B}}}

\newcommand{\ctxset}{\ensuremath{\mathcal{C}}}

\newcommand{\dbirth}{\textsc{Birth}}
\newcommand{\ddeepen}{\textsc{Deepening}}
\newcommand{\devolve}{\textsc{Evolve}}

\newcommand{\intraedges}{intra-topic transition edges}
\newcommand{\interedges}{inter-topic transition edges}

\newcommand{\chrono}{\textsc{ChronoExpand}}

\title{\method{}: Transition-Aware Retrieval for Evolving Personal Memory}

\author{%
  Shigeng Chen\textsuperscript{1}\thanks{Equal contribution.}\quad
  Linhao Luo\textsuperscript{2}\footnotemark[1]\quad
  Changlong Shi\textsuperscript{3}\quad
  Zhangchi Qiu\textsuperscript{1}\quad
  Shirui Pan\textsuperscript{1}\thanks{Corresponding author.} \\
  \textsuperscript{1}Griffith University\quad
  \textsuperscript{2}Monash University \\
  \textsuperscript{3}The Hong Kong Polytechnic University \\
  \texttt{shigeng.chen@griffithuni.edu.au}\quad
  \texttt{s.pan@griffith.edu.au}
}

\begin{document}

\maketitle

\begin{abstract}
Long-term personalized agents are increasingly expected to answer queries over user histories in which preferences, goals, and constraints evolve across interleaved conversations.
Existing retrieval-based and structured-memory systems can surface relevant facts, but often return isolated memory units that do not explain which earlier preference was refined, contradicted, or superseded by later evidence.
We introduce \method{}\footnote{Code is available at \url{https://github.com/Aries-chen/TransMem}.}, a transition-aware memory retrieval framework that stores preference updates as explicit retrieval units alongside ordinary factual memories.
During memory construction, \method{} routes user interaction history into topic-structured memory graphs, detects shift events, and grounds them as transition records.
At inference time, \method{} first retrieves factual anchors and then adaptively expands over transition records, producing a compact context of factual anchors and accepted transition records.
Experiments on PersonaMem and PersonaBench across two Qwen backbones show that \method{} outperforms representative retrieval-based and structured-memory baselines.
Controlled analyses further show that transition records and adaptive expansion contribute beyond factual anchoring, chronological expansion, flat transition retrieval, and fixed-hop traversal.
These results suggest that preference-update records are useful retrieval units for long-term personalization.

\end{abstract}

\section{Introduction}
\label{sec:introduction}

Long-term memory is becoming central to personalized language agents. As LLM-based assistants interact with users over extended periods, they must track not only isolated facts but also evolving preferences, goals, routines, and constraints~\citep{generativeagents2023,memorybank2024,memgpt2023}. This is difficult because users are non-stationary: a preference for quick solo workouts may shift toward longer, social outdoor activities after joining a hiking group. These changes are rarely cleanly segmented; outdated preferences, transient states, and current intentions often coexist across interleaved conversational threads. Multi-turn fragility further compounds the problem, as even strong LLMs can over-rely on premature assumptions and fail to recover from early errors~\citep{lostmultiturn2026}. Recent benchmarks similarly show persistent weaknesses in multi-session reasoning, temporal updating, and dynamic user-profile tracking under long, topic-interleaved contexts~\citep{lee2025realtalk,locomo2024,longmemeval2025,personabench,personamem2025}.

\begin{figure}[!t]
  \centering
  \includegraphics[width=0.94\linewidth]{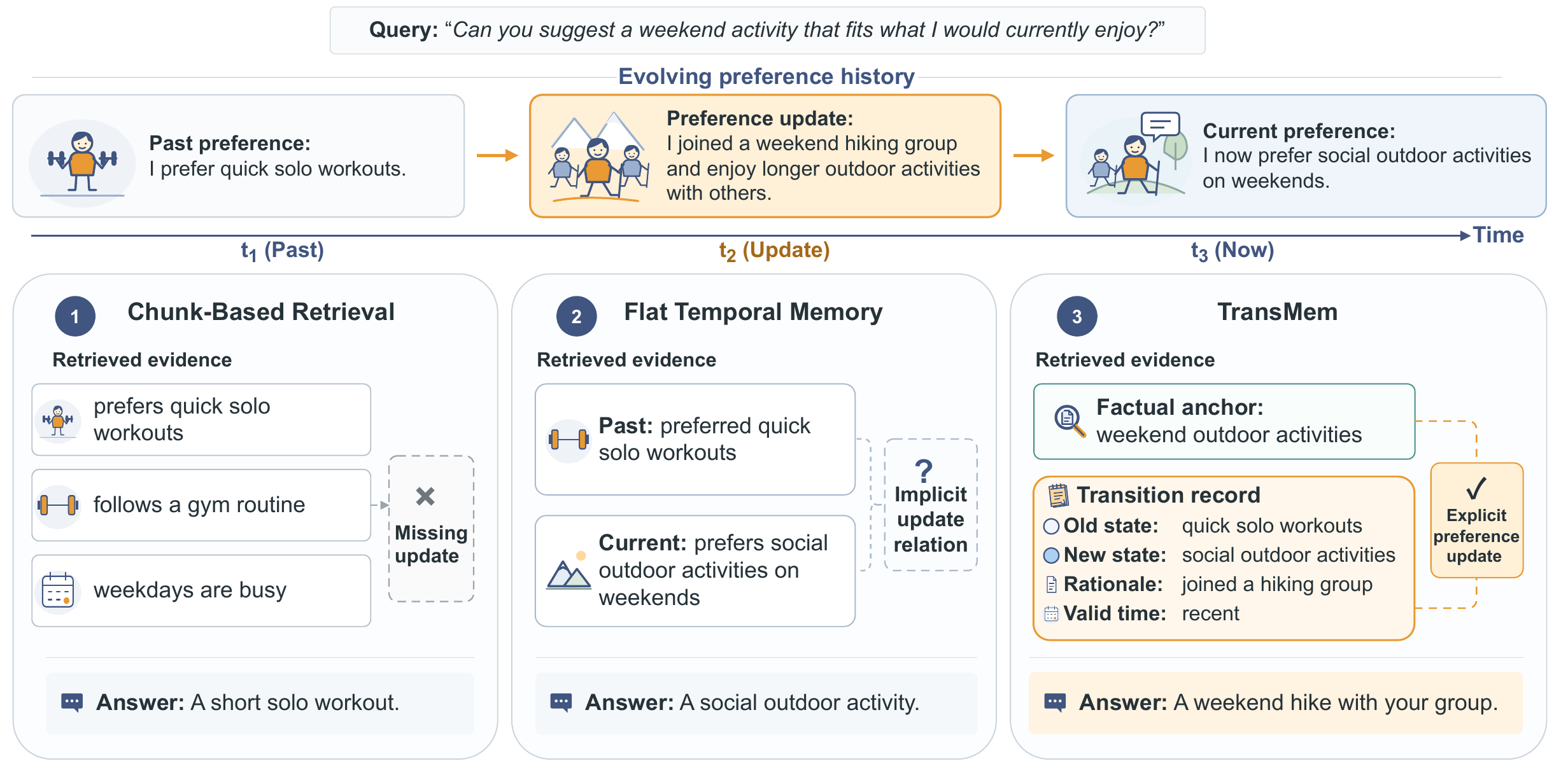}
  \caption{Evolving personal preferences: snippets can be stale and temporal facts leave updates implicit, while \method{} retrieves factual anchors with explicit transitions.}
  \label{fig:motivation}
\end{figure}

Existing memory-augmented approaches have made substantial progress in storing, organizing, and retrieving personal information. Traditional retrieval-augmented generation (RAG) retrieves semantically similar chunks from past interactions~\citep{rag2020}, but such similarity-driven retrieval can surface stale or noisy evidence when user histories span long time periods, especially under knowledge volatility and temporal conflicts~\citep{temporalsurvey2025}. Recent memory systems introduce stronger organization and consolidation mechanisms, including sentence graph memory~\citep{SGMem2026}, temporal knowledge graphs~\citep{zep2025}, scalable memory stores~\citep{mem02025}, agentic note organization~\citep{amem2025}, staged memory architectures~\citep{lightmem2026}, and recollection-style retrieval~\citep{rfmem2026}. These systems improve how memories are stored or selected, yet evolving personal preferences introduce a distinct retrieval challenge: the decisive evidence is often not a single memory item, but the update that explains how one user state superseded another.

A subtle difficulty is that the evidence needed to resolve a preference update is not always the evidence most similar to the query. A user may ask for a current recommendation without mentioning the historical event that changed the preference. Even when old and new facts are retrieved together, presenting them as independent snippets leaves the preference-update relation for the model to infer. The useful evidence is therefore often the update itself: what the user previously preferred, what replaced it, why it changed, and when the new state became valid.

Figure~\ref{fig:motivation} contrasts recommendations based on stale snippets, independent temporal facts, and explicit preference-update evidence.
This motivates \emph{preference-transition retrieval}: retrieving the update relation itself rather than leaving the model to infer it from disconnected factual snippets.
We introduce \textbf{\method{}}, a transition-aware memory retrieval framework that stores preference updates as transition records alongside ordinary factual memories.
During construction, \method{} routes streaming interactions into topic-structured memory graphs and detects shift events as transition records.
During retrieval, it first grounds the query in factual anchors and then adaptively expands over adjacent transition records when update evidence is needed.
This design recovers compact preference-update trajectories while avoiding unrelated chronological neighbors and keeping online retrieval lightweight.

We evaluate \method{} on complementary personalized-memory benchmarks, with PersonaMem~\citep{personamem2025} stressing long-context, topic-interleaved preference evolution and PersonaBench~\citep{personabench} testing broader personalized QA.
Experimental results show that \method{} consistently achieves the best overall score among representative baselines.
Ablation and control experiments with factual-only retrieval, chronological expansion, flat transition retrieval, and fixed-hop expansion show that the gains are not explained by factual anchoring, temporal-neighbor context, direct transition-text retrieval, or fixed traversal depth alone.

Our contributions are as follows.
First, we identify preference updates as a missing retrieval unit in long-term personalized memory and represent them as structured transition records.
Second, we introduce \method{}, which combines topic-structured memory construction, shift-event detection, and adaptive transition-aware retrieval.
Third, we evaluate \method{} on PersonaMem and PersonaBench across two Qwen backbones, with controlled analyses isolating the roles of transition records and adaptive expansion.

\clearpage
\begin{figure}[!t]
  \centering
  \includegraphics[width=\linewidth]{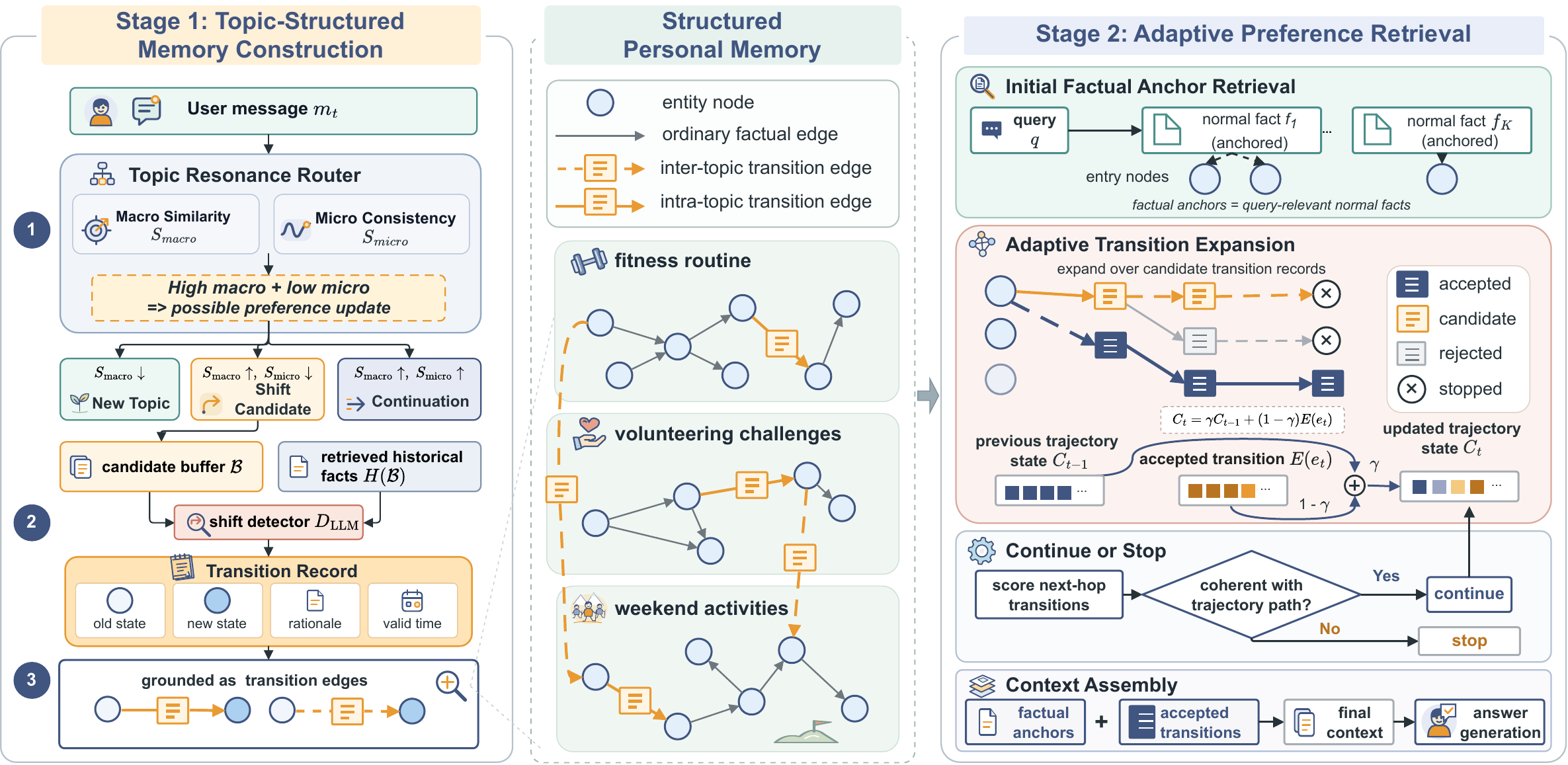}
  \caption{Overview of \method{}.
Stage 1 (memory construction): \textcolor[HTML]{405580}{\ding{182}} organize messages into topic graphs via the Topic Resonance Router; \textcolor[HTML]{405580}{\ding{183}} compare candidate messages with historical facts to extract transition records; \textcolor[HTML]{405580}{\ding{184}} ground transition records into structured personal memory shown in the center.
Stage 2 (inference): retrieve factual anchors and adaptively expand over transition records to assemble the final context.}
  \label{fig:overview}
\end{figure}

\section{Methodology}
\label{sec:methodology}

\paragraph{Preliminary.}
We consider personalized question answering for a user $u$ with chronological message history $(m_1,\ldots,m_t)$.
A memory-augmented system builds a memory $\mathcal{M}_u$ from this history and retrieves an evidence context $\ctxset(q)$ for a query $q$.
An LLM with parameters $\theta$ then generates an answer $y$ conditioned on the query and retrieved evidence:
\[
\ctxset(q)=\operatorname{Retrieve}(q,\mathcal{M}_u),
\qquad
y\sim p_\theta\!\left(\cdot\mid q,\ctxset(q)\right).
\]
Questions may concern historical facts, current preferences, or changes across time.

\paragraph{Overview.}
We propose \method{}, a two-stage framework for retrieving evidence from long-term personal memory under evolving user preferences.
The construction stage organizes each user's interaction history into topic-structured memory graphs and stores both ordinary factual memories and explicit preference-update records.
The retrieval stage first selects query-relevant factual anchors and then adaptively expands over stored transition records when preference-update evidence is needed.
Figure~\ref{fig:overview} illustrates the overall pipeline.

\paragraph{Notation.}
In \method{}, the memory $\mathcal{M}_u$ is represented by the user memory graph $G_u$.
We use \emph{topic} to denote a fine-grained, user-specific topic graph (e.g., fitness routine), rather than a coarse domain label such as hobbies or work.
We denote the $i$-th topic graph by $T_i$ with summary embedding $s_i$.
A \emph{normal fact} is an atomic, non-transition memory extracted from the user history.
A \emph{transition record} is a structured old-to-new preference update detected from historical evidence.
A transition record stores the content of the update, while its grounded transition edge provides the graph structure used for traversal.

\subsection{Topic-Structured Memory Construction}
\label{sec:construction}

\paragraph{Topic Resonance Router.}
Each incoming message $m_t$ is routed through a two-stage decision over existing topic graphs.
The first gate measures semantic alignment with each topic summary,
\begin{equation}
S_{\mathrm{macro}}(m_t, T_i)
=
\cos\!\left(E(m_t), s_i\right),
\label{eq:macro}
\end{equation}
where $E(\cdot)$ is the text embedding operator.
Let
\begin{equation}
i^* = \operatorname*{arg\,max}_i S_{\mathrm{macro}}(m_t,T_i),
\qquad
M_t = S_{\mathrm{macro}}(m_t,T_{i^*}).
\end{equation}
The second gate measures entity-level consistency with the candidate topic, namely how strongly the entities in $m_t$ overlap with the entities already associated with $T_{i^*}$:
\begin{equation}
S_{\mathrm{micro}}(m_t, T_{i^*})
=
\frac{|K(m_t) \cap K(T_{i^*})|}
{\max(1, |K(m_t)|)},
\label{eq:micro}
\end{equation}
where $K(m_t)$ extracts entities from the message and $K(T_i)$ denotes entities already associated with facts in topic graph $T_i$.
The denominator is clipped by one so that messages without extracted entities do not produce undefined scores.
We write $\mu_t = S_{\mathrm{micro}}(m_t, T_{i^*})$.

The routing decision is
\begin{equation}
r_t =
\begin{cases}
\textsc{Birth}, & M_t < \tau_{\mathrm{macro}},\\
\textsc{Deepening}, & M_t \ge \tau_{\mathrm{macro}} \wedge \mu_t \ge \tau_{\mathrm{micro}},\\
\textsc{Evolve}, & M_t \ge \tau_{\mathrm{macro}} \wedge \mu_t < \tau_{\mathrm{micro}}.
\end{cases}
\label{eq:routing}
\end{equation}
A low macro score creates a new topic graph.
A high macro score and high micro score indicate that the message continues an existing topic.
A high macro score but low micro score means that the message is topically related but introduces different entities or constraints; \method{} therefore treats it as a candidate preference update rather than immediately committing it as a transition.
Regardless of the routing outcome, each message is ingested into $G_u$ so that the memory graph remains a complete record.

\paragraph{Shift Detection.}
Messages routed as \textsc{Evolve} are accumulated in a candidate-update buffer $\mathcal{B}$.
Detection is triggered only when the buffer reaches size $\kappa$, which reduces spurious updates caused by isolated or ambiguous messages.
The router determines which messages become candidate updates, while the detector compares the resulting buffer against retrieved historical facts from the user memory graph:
\begin{equation}
\mathcal{H}(\mathcal{B})
=
\operatorname*{arg\,topK}_{f \in \mathcal{F}_u}
s_{\mathrm{hist}}(\mathcal{B}, f),
\label{eq:hist}
\end{equation}
where $\mathcal{F}_u$ denotes the normal facts stored in $\usergraph$, and $s_{\mathrm{hist}}$ scores their relevance to the buffered candidate messages.

An LLM-based shift detector then compares the buffered candidate messages with the retrieved historical facts and outputs structured shift events:
\begin{equation}
\mathcal{Z}
=
\mathcal{D}_{\mathrm{LLM}}\!\left(\mathcal{B}, \mathcal{H}(\mathcal{B})\right).
\label{eq:detector}
\end{equation}
We use \emph{shift event} for the detector output and \emph{transition record} for the stored retrieval unit derived from it.
Each detected update is represented as
\begin{equation}
z =
\big(s^{-}, s^{+}, r, g, t_{\mathrm{valid}}\big),
\label{eq:shift}
\end{equation}
where $s^{-}$ is the old preference state, $s^{+}$ is the new preference state, $r$ is the event type, $g$ is the update rationale explaining why the change occurred, and $t_{\mathrm{valid}}$ records when the new state becomes valid.

\paragraph{Transition Linking and Edge Construction.}
Let $V_u$ be the set of entity nodes in $G_u$, where entities include concepts such as activities and interests.
Each topic graph $T_i$ is a subgraph of $G_u$.
For each detected update $z \in \mathcal{Z}$, we ground its old and new states to existing nodes:
\begin{equation}
v^{a} = \operatorname*{arg\,max}_{v \in V_u}
\cos\!\left(E(s^{a}), E(v)\right),
\qquad a \in \{-,+\}.
\label{eq:grounding}
\end{equation}
Here, $E(v)$ denotes the text embedding of the node's entity or concept name.
We retain the update only if both endpoint similarities are at least $\tau_g$.
The existing nodes provide structural anchors, while the transition record preserves the complete states $s^{-}$ and $s^{+}$.
Each retained record is stored with a directed transition edge:
\begin{equation}
e_z =
\big(v^{-}, v^{+}, r, y_e, g, t_{\mathrm{valid}}\big),
\qquad
y_e \in \{\mathrm{intra}, \mathrm{inter}\},
\label{eq:transition-edge}
\end{equation}
where $v^{-}$ and $v^{+}$ are the grounded endpoints.
The stored text follows the detector fields and is embedded as the retrieval representation of the transition record.
The same shift detector is used for all transitions.
After grounding $s^{-}$ and $s^{+}$ to graph nodes, we label the resulting transition according to endpoint assignment:
it is an \emph{intra-topic transition edge} if both endpoints belong to the same topic graph, and an \emph{inter-topic transition edge} otherwise.
Thus, the intra/inter distinction reflects where the old and new states are grounded, not a separate detector or supervision signal.
The update rationale $g$ and valid time $t_{\mathrm{valid}}$ are retained as part of the transition record and may be verbalized in the final retrieved context.

\subsection{Adaptive Preference Retrieval}
\label{sec:retrieval}

At inference time, \method{} retrieves evidence in two stages: initial factual anchoring, followed by adaptive transition expansion.
The final context contains factual anchors and accepted transition records, whose fields include the old and new states.

\paragraph{Initial Factual Anchor Retrieval.}
Given a query $q$, \method{} first retrieves query-relevant normal facts:
\begin{equation}
\mathcal{A}(q)
=
\operatorname*{arg\,topK}_{f \in \mathcal{F}_{\mathrm{normal}}}
s(q,f),
\label{eq:anchor}
\end{equation}
where $\mathcal{F}_{\mathrm{normal}}$ contains all normal factual memories and $s(q,f)$ is the factual relevance score.
The selected facts $\mathcal{A}(q)$ are \emph{factual anchors}: they are included directly in the final context and also provide entry points into the transition graph.
Each normal fact is stored as an atomic fact edge in $G_u$ with head and tail nodes; the endpoints of selected factual anchors provide the entry nodes for transition expansion.

\paragraph{Adaptive Transition Expansion.}
Let $p$ denote a partial transition path.
For the empty path, candidate transition edges are adjacent to the endpoints of the factual anchors; after a transition is accepted, candidates are adjacent to the target endpoint of the current path.
For retrieval scoring, each transition edge is represented by the text of its associated transition record.
Let $C_p$ be the trajectory state of path $p$, with $C_{\emptyset}=E(q)$.
A candidate next transition $e$ is scored by
\begin{equation}
\sigma(e \mid p)
=
\cos\!\left(C_p, E(e)\right),
\label{eq:path-score}
\end{equation}
where $E(e)$ embeds the text associated with transition edge $e$.
A candidate is retained only if it remains coherent with the current trajectory:
\begin{equation}
e \in \Phi(p)
\quad \Longleftrightarrow \quad
\sigma(e \mid p) \ge \tau_c
\;\wedge\;
\left(
p=\emptyset
\;\vee\;
\sigma(e \mid p)-\sigma_{\mathrm{last}}(p) \ge -\delta_c
\right),
\label{eq:path-continue}
\end{equation}
where $\tau_c$ is the continuation threshold, $\delta_c$ is the drop tolerance, and $\sigma_{\mathrm{last}}(p)$ is the score of the most recently accepted transition in $p$.
The first condition requires the next transition to remain relevant to the evolving trajectory.
The second condition prevents abrupt coherence collapse within a single hop.

When a transition edge $e$ is accepted, its target endpoint determines the next structural neighborhood, while its associated transition record updates the trajectory state:
\begin{equation}
C_{p \oplus e}
=
\gamma C_p + (1-\gamma)E(e),
\label{eq:state-update}
\end{equation}
where $p \oplus e$ denotes the path obtained by appending $e$.
This interpolation keeps the trajectory state anchored to the original query while allowing accepted transition records to steer later hops, preventing expansion from drifting to locally adjacent but semantically unrelated transitions.
The final retrieved context consists of factual anchors and accepted transition records, allowing the QA model to resolve stale and current preferences without retrieving unrelated chronological neighbors.

\section{Experiments}
\label{sec:experiments}

\subsection{Experimental Setup}

\paragraph{Datasets.}
We evaluate on two complementary personalized-memory benchmarks.
\textbf{PersonaMem}~\citep{personamem2025} directly stresses long-context preference evolution: each user is associated with a chronological message stream and questions spanning seven preference- and memory-related categories, and we use both its 32k-token and 128k-token context regimes.
\textbf{PersonaBench}~\citep{personabench} provides a broader personalized-QA setting over synthetic private user data and evaluates answer quality with lexical F1.
Together, the two benchmarks test whether transition-aware retrieval helps both in long, topic-interleaved preference histories and in a separate personalized-QA benchmark.
Dataset and evaluation details are provided in Appendix~\ref{app:datasets}.

\paragraph{Baselines.}
We compare \method{} against three baseline groups.
\emph{Zeroshot-only}: \textsc{ZeroMem} provides no user history.
\emph{Retrieval-based}: \textsc{NaiveRAG}~\citep{rag2020} and \textsc{RF-Mem}~\citep{rfmem2026} select passages by query relevance.
\emph{Structured memory}: \textsc{Zep}~\citep{zep2025}, \textsc{A-Mem}~\citep{amem2025}, \textsc{Mem0}~\citep{mem02025}, and \textsc{LightMem}~\citep{lightmem2026} maintain external user memory representations.

\paragraph{Implementation Settings.}
We use two open-source backbones, Qwen3-8B~\citep{qwen3_2025} and Qwen3.5-27B~\citep{qwen35_2026}, with each used for both memory construction and final answer generation.
Implementation details are reported in Appendix~\ref{app:impl-details}.

\subsection{Main Results}

We first test whether retrieving preference-update records improves end-to-end personalized QA.
Across the evaluated PersonaMem and PersonaBench settings, \method{} achieves the highest overall score among representative retrieval-based and structured-memory baselines, with the largest improvement reaching 4.84 overall points.
Controlled analyses with factual-only retrieval, chronological expansion, flat transition retrieval, and fixed-hop expansion further show that the improvements are not explained by factual anchoring, temporal-neighbor context, direct transition-text retrieval, or fixed traversal depth alone.

\begin{table}[!t]
\centering
\caption{PersonaMem results in the 32k context regime. Best results in bold; second-best underlined.}
\label{tab:main-32k}
\footnotesize
\setlength{\tabcolsep}{3.5pt}
\renewcommand{\arraystretch}{1.00}
\begin{tabular}{l l c c c c c c c >{\columncolor{gray!12}}c}
\toprule
& Method & \makecell{Revisit\\Reasons} & \makecell{Track\\Evolution} & \makecell{Latest\\Prefs} & \makecell{Aligned\\Recs} & \makecell{New\\Scenario} & \makecell{Shared\\Facts} & \makecell{New\\Ideas} & Overall \\
\midrule
\multirow{8}{*}{\rotatebox[origin=c]{90}{\textit{Qwen3-8B}}}
& \textsc{ZeroMem}   & 57.41 & 55.00 & 29.41 & 29.63 & 18.37 & 35.40 & \phantom{0}9.09 & 34.53 \\
& \textsc{NaiveRAG}  & \underline{68.52} & 54.00 & 58.82 & 38.89 & 51.02 & 54.87 & 18.18 & 47.37 \\
& \textsc{RF-Mem}    & 53.70 & 56.00 & 35.29 & 61.11 & 42.86 & 57.52 & 19.32 & 47.79 \\
& \textsc{Zep}       & \underline{68.52} & 55.00 & 52.94 & \textbf{66.67} & 61.22 & 54.87 & 22.73 & 52.42 \\
& \textsc{A-Mem}     & \textbf{75.93} & \underline{61.00} & \textbf{70.59} & \underline{62.96} & 44.90 & 53.98 & \underline{23.86} & \underline{53.05} \\
& \textsc{Mem0}      & 66.67 & 51.00 & \underline{64.71} & 61.11 & 48.98 & 51.33 & 10.23 & 46.74 \\
& \textsc{LightMem}  & 59.26 & 50.00 & \underline{64.71} & 61.11 & \underline{63.27} & \textbf{63.72} & 20.45 & 52.00 \\
& \method{}~(ours)   & \textbf{75.93} & \textbf{62.00} & 58.82 & 59.26 & \textbf{67.35} & \underline{61.06} & \textbf{26.14} & \textbf{56.84} \\
\cmidrule(lr){2-10}
\multirow{8}{*}{\rotatebox[origin=c]{90}{\textit{Qwen3.5-27B}}}
& \textsc{ZeroMem}   & 66.67 & 57.00 & 23.53 & 42.59 & 26.53 & 36.28 & \phantom{0}7.95 & 38.11 \\
& \textsc{NaiveRAG}  & 75.93 & \underline{64.00} & 64.71 & 70.37 & 71.43 & 61.06 & 34.09 & 60.63 \\
& \textsc{RF-Mem}    & \textbf{85.19} & 55.00 & 64.71 & \underline{77.78} & 65.31 & 61.06 & 40.91 & 61.26 \\
& \textsc{Zep}       & 79.63 & 61.00 & \underline{70.59} & 72.22 & 81.63 & 62.83 & 40.91 & 63.58 \\
& \textsc{A-Mem}     & 68.52 & 61.00 & \underline{70.59} & 75.93 & 61.22 & \underline{71.68} & \textbf{50.00} & \underline{64.42} \\
& \textsc{Mem0}      & \underline{81.48} & \textbf{65.00} & 47.06 & 74.07 & \underline{87.76} & \textbf{72.57} & 25.00 & 64.00 \\
& \textsc{LightMem}  & 66.67 & 61.00 & 64.71 & 70.37 & 71.43 & 65.49 & 35.23 & 60.21 \\
& \method{}~(ours)   & \textbf{85.19} & 62.00 & \textbf{82.35} & \textbf{79.63} & \textbf{89.80} & \underline{71.68} & \underline{44.32} & \textbf{69.26} \\
\bottomrule
\end{tabular}
\par\medskip
\normalsize
\caption{PersonaMem results in the 128k context regime. Best results in bold; second-best underlined.}
\label{tab:main-128k}
\footnotesize
\setlength{\tabcolsep}{3.5pt}
\renewcommand{\arraystretch}{1.00}
\begin{tabular}{l l c c c c c c c >{\columncolor{gray!12}}c}
\toprule
& Method & \makecell{Revisit\\Reasons} & \makecell{Track\\Evolution} & \makecell{Latest\\Prefs} & \makecell{Aligned\\Recs} & \makecell{New\\Scenario} & \makecell{Shared\\Facts} & \makecell{New\\Ideas} & Overall \\
\midrule
\multirow{8}{*}{\rotatebox[origin=c]{90}{\textit{Qwen3-8B}}}
& \textsc{ZeroMem}   & 44.30 & 50.91 & 34.83 & 30.65 & 38.60 & 26.62 & 17.86 & 32.67 \\
& \textsc{NaiveRAG}  & 48.10 & 53.18 & \underline{56.60} & 46.13 & 32.16 & 53.25 & 21.13 & 44.42 \\
& \textsc{RF-Mem}    & 46.84 & \underline{54.55} & 52.83 & 43.23 & 28.65 & 48.05 & 22.22 & 42.43 \\
& \textsc{Zep}       & 50.63 & 52.73 & 54.28 & 49.03 & 45.61 & 58.44 & 22.22 & 45.90 \\
& \textsc{A-Mem}     & \underline{51.27} & 53.64 & 52.54 & \textbf{53.55} & 29.82 & 51.95 & \textbf{26.14} & 45.26 \\
& \textsc{Mem0}      & 50.00 & 53.64 & 56.17 & 45.48 & 32.75 & 51.30 & 21.35 & 44.33 \\
& \textsc{LightMem}  & 44.94 & 52.73 & 55.30 & 46.13 & \underline{47.95} & \textbf{64.29} & \underline{23.53} & \underline{46.27} \\
& \method{}~(ours)   & \textbf{51.90} & \textbf{55.00} & \textbf{56.75} & \underline{51.29} & \textbf{53.22} & \underline{62.99} & 21.13 & \textbf{48.03} \\
\cmidrule(lr){2-10}
\multirow{8}{*}{\rotatebox[origin=c]{90}{\textit{Qwen3.5-27B}}}
& \textsc{ZeroMem}   & 63.92 & 58.64 & 38.90 & 32.90 & 31.58 & 47.40 & 32.24 & 40.49 \\
& \textsc{NaiveRAG}  & \textbf{72.78} & 59.55 & 63.43 & 62.26 & 47.37 & 79.22 & 41.61 & 58.77 \\
& \textsc{RF-Mem}    & \underline{72.15} & \underline{60.91} & \underline{69.81} & 58.06 & 50.29 & 78.57 & 38.56 & 59.83 \\
& \textsc{Zep}       & 69.62 & 50.45 & \textbf{70.54} & 63.23 & 51.46 & 73.38 & \underline{43.36} & 60.30 \\
& \textsc{A-Mem}     & 68.35 & 60.45 & 66.04 & \underline{66.13} & \underline{68.42} & 81.17 & 41.18 & \underline{61.64} \\
& \textsc{Mem0}      & \textbf{72.78} & 59.55 & 64.88 & 62.26 & \underline{68.42} & \underline{81.82} & 40.52 & 60.85 \\
& \textsc{LightMem}  & 67.72 & 57.73 & 62.41 & 55.16 & 57.31 & 79.22 & 41.39 & 57.61 \\
& \method{}~(ours)   & 70.89 & \textbf{62.27} & 69.23 & \textbf{67.74} & \textbf{73.68} & \textbf{85.06} & \textbf{44.66} & \textbf{64.69} \\
\bottomrule
\end{tabular}
\end{table}

Table~\ref{tab:main-32k} shows that \method{} does not dominate every category, but it consistently improves the overall score under both backbones.
Baselines remain competitive on locally answerable or stable-fact categories, while \method{} achieves the highest \emph{New Scenario} accuracy under both backbones and the highest \emph{Track Evolution} accuracy with Qwen3-8B.
This pattern is consistent with our view that the useful retrieval unit is often not an isolated fact but the transition that explains how one user state supersedes another.

Table~\ref{tab:main-128k} reports the 128k regime, which more directly stresses the failure mode in Figure~\ref{fig:motivation}: relevant evidence is separated by longer spans of unrelated history, and old and new user states are more likely to coexist.
Under this regime, \method{} again achieves the best overall accuracy with both backbones.
The category-level pattern is most informative on \emph{Track Evolution} and \emph{New Scenario}, where \method{} is strongest under both backbones.
Open-ended ideation categories such as \emph{New Ideas} are less diagnostic of transition retrieval because success also depends on preference-consistent generation and creative extrapolation; we analyze one such case in Appendix~\ref{app:case-analysis}.
We further note that with Qwen3.5-27B, \textsc{FullText} attains the strongest scores on \emph{Revisit Reasons} and \emph{Track Evolution}, where direct access to the complete history removes retrieval noise; this advantage scales linearly with context cost and is unavailable for Qwen3-8B in the 128k regime (Appendix~\ref{app:fulltext}).

\begin{table}[!t]
\centering
\caption{PersonaBench results (lexical F1, \%). Best results in bold; second-best underlined.}
\label{tab:main-personabench}
\footnotesize
\setlength{\tabcolsep}{4pt}
\renewcommand{\arraystretch}{1.05}
\begin{tabular}{l l c c c c >{\columncolor{gray!12}}c}
\toprule
& Method & Basic Info & Pref Easy & Pref Hard & Social & Overall \\
\midrule
\multirow{5}{*}{\rotatebox[origin=c]{90}{\textit{Qwen3-8B}}}
& \textsc{NaiveRAG}  & 13.96 & 21.29 & 24.48 & 15.65 & 17.88 \\
& \textsc{RF-Mem}    & 18.63 & \underline{23.80} & 18.60 & 15.88 & 18.58 \\
& \textsc{A-Mem}     & 16.71 & 21.54 & 26.71 & 21.82 & 20.21 \\
& \textsc{LightMem}  & \underline{19.37} & 23.38 & \underline{29.05} & \textbf{23.21} & \underline{22.43} \\
& \method{}~(ours)   & \textbf{23.80} & \textbf{25.14} & \textbf{30.16} & \underline{22.00} & \textbf{24.67} \\
\cmidrule(lr){2-7}
\multirow{5}{*}{\rotatebox[origin=c]{90}{\textit{Qwen3.5-27B}}}
& \textsc{NaiveRAG}  & 14.31 & 23.11 & 22.93 & 16.08 & 18.07 \\
& \textsc{RF-Mem}    & 14.61 & 16.08 & 23.11 & 22.93 & 18.21 \\
& \textsc{A-Mem}     & 27.34 & 23.63 & \underline{24.53} & \underline{35.85} & \underline{28.38} \\
& \textsc{LightMem}  & \textbf{33.26} & \underline{31.61} & 23.29 & 20.45 & 28.34 \\
& \method{}~(ours)   & \underline{28.19} & \textbf{31.79} & \textbf{24.95} & \textbf{38.46} & \textbf{30.38} \\
\bottomrule
\end{tabular}
\end{table}


PersonaBench provides a complementary open-ended evaluation beyond PersonaMem's multiple-choice format. 
As shown in Table~\ref{tab:main-personabench}, \method{} achieves the best overall F1 under both backbones, with category-level differences reflecting the open-ended nature of this benchmark. 
These results indicate that transition-aware memory remains useful beyond PersonaMem, although absolute F1 remains modest in this harder open-ended setting.


\par\medskip
\noindent\begin{minipage}{\linewidth}
\centering
\includegraphics[width=0.92\linewidth]{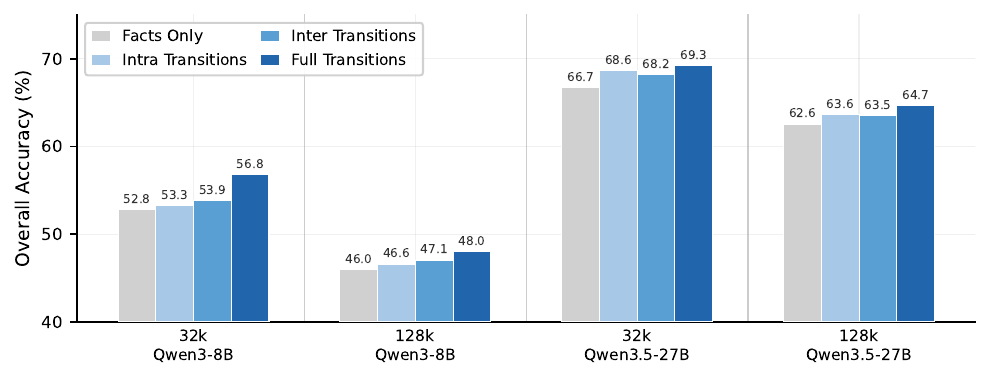}
\captionof{figure}{Ablation results comparing four retrieval variants across backbone and context settings.
\textit{Facts Only}: factual retrieval without transition edges;
\textit{Intra Transitions}: intra-topic transition edges only;
\textit{Inter Transitions}: inter-topic transition edges only;
\textit{Full Transitions}: \method{} with both edge types.}
\label{fig:ablation}
\end{minipage}
\par\nopagebreak[4]

\subsection{Ablation Study}
\label{sec:analysis}

We next isolate whether explicit transition structure contributes beyond factual anchoring.
Figure~\ref{fig:ablation} compares four retrieval variants that progressively add transition-edge structure across both backbones and context regimes.
Across all four evaluation settings, the \emph{Facts Only} variant yields the lowest overall accuracy, indicating that time-aware factual anchors alone do not fully resolve revised or conflicting user states.
Adding either intra-topic or inter-topic transition edges improves over the factual-only baseline, while the full model performs best in every setting.
In three of four settings, inter-topic transition edges contribute slightly more than intra-topic edges alone (e.g., +1.05 vs.\ +0.42\,pp on 32k-8B), though the 32k-27B setting reverses this order (+1.89 vs.\ +1.47\,pp).
Thus, transition records provide information beyond factual anchors, and intra- and inter-topic transitions contribute complementary signals.
A per-category comparison is shown in Figure~\ref{fig:per-category-8b}.

\subsection{Accuracy--Efficiency Trade-off}
\label{sec:efficiency}

\begin{figure}[t]
\centering
\begin{minipage}[t]{0.46\linewidth}
  \centering
  \includegraphics[width=\linewidth]{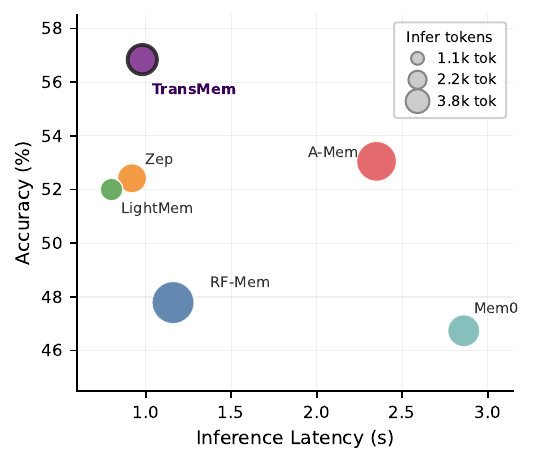}
\end{minipage}
\hspace{0.02\linewidth}
\begin{minipage}[t]{0.46\linewidth}
  \centering
  \includegraphics[width=\linewidth]{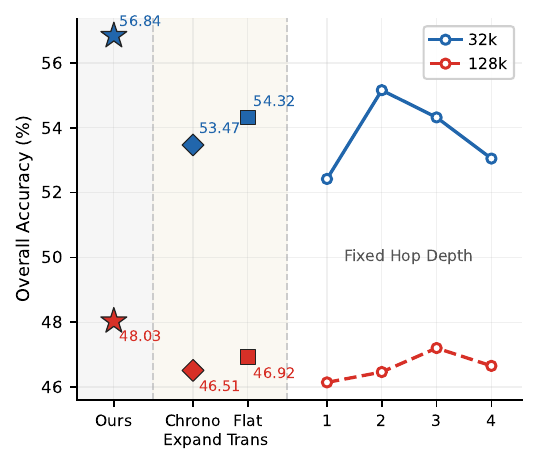}
\end{minipage}
\caption{
\textbf{Left}: Inference latency and token usage versus overall accuracy in the 32k-8B setting; bubble area is proportional to inference token count.
\textbf{Right}: Comparison of \method{} with chronological expansion (\chrono{}), flat transition retrieval (\textsc{FlatTrans}), and fixed-hop transition expansion.
}
\label{fig:eff-hops}
\end{figure}



We then ask whether the improvements come from retrieving more evidence or more informative evidence.
We measure end-to-end inference time from query submission to final answer, including retrieval, context assembly, retrieval-time model calls, and QA generation.
In our reproduced pipelines, \textsc{Mem0} and \textsc{A-Mem} invoke an additional LLM call during retrieval, whereas \method{} performs embedding-space expansion after memory construction.
As shown in Figure~\ref{fig:eff-hops} (left), \method{} achieves the highest accuracy with 1.82k inference tokens and 0.98\,s latency, compared with 3.4k tokens and 2.35\,s for the next-best system by accuracy, \textsc{A-Mem}.
This suggests that the improvements come from retrieving more targeted evidence rather than simply increasing the amount of context.

\subsection{Analysis of Retrieval Expansion Strategies}
\label{sec:hops}


Finally, we isolate the expansion mechanism using three controls from the same factual anchors.
\textsc{ChronoExpand} tests whether ordinary temporal neighbors are sufficient by adding chronological normal-fact neighbors from the same topic timeline.
\textsc{FlatTrans} tests whether transition records alone are sufficient by retrieving them as flat documents with query--transition similarity, without following graph endpoints.
Fixed-hop variants test whether a manually chosen traversal depth can replace adaptive stopping.
\method{} differs from all three by following graph-grounded transition endpoints while updating a trajectory state during expansion.
These controls correspond to the main retrieval choices in Figure~\ref{fig:overview}: whether to use transition records, whether to follow grounded endpoints, and whether to stop adaptively.
Further details are in Appendix~\ref{app:retrieval-strategies}.

Figure~\ref{fig:eff-hops} (right) compares these strategies for the Qwen3-8B backbone under both context regimes.
\textsc{ChronoExpand} (53.47\% / 46.51\%) improves over the worst fixed depths but trails \method{}, showing that nearby chronological normal facts alone are not enough.
\textsc{FlatTrans} (54.32\% / 46.92\%) improves further, confirming that transition records carry useful signal, but still falls short of \method{} (56.84\% / 48.03\%).
This indicates that the gain comes not only from storing transition text, but also from using the memory graph to choose which transition records to consider next.
Fixed-hop retrieval further shows that no single depth is uniformly appropriate: hop~2 is best in 32k while hop~3 is best in 128k, whereas adaptive expansion exceeds the best fixed alternative in both regimes (by 1.68\,pp and 0.83\,pp respectively).
Together, these controls show that transition records are a useful retrieval unit, but their strongest effect comes from using them in an adaptive graph expansion rather than retrieving them as isolated text.

\par\medskip
\noindent\begin{minipage}{\linewidth}
\centering
\includegraphics[width=0.80\linewidth]{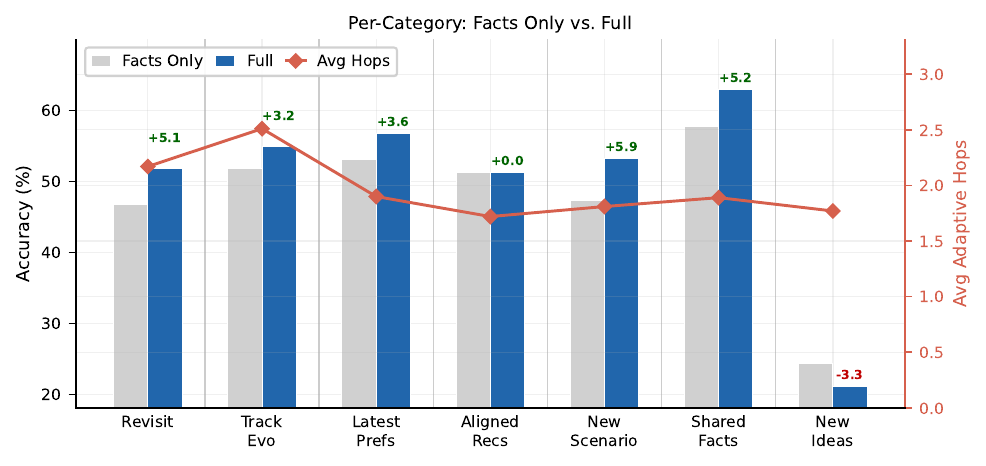}
\captionof{figure}{Per-category accuracy and adaptive retrieval depth on PersonaMem-128k (Qwen3-8B). Bars compare \textit{Facts Only} with \textit{Full Transitions}; the right axis shows average expansion depth.}
\label{fig:per-category-8b}
\end{minipage}
\par\nopagebreak[4]

\paragraph{Category-wise Gains and Adaptive Depth.}
Figure~\ref{fig:per-category-8b} examines how these gains vary across question categories in the 128k-8B setting.
Full transition retrieval improves accuracy in five of seven categories, with the largest gains on \emph{New Scenario} ($+5.85$\,pp), \emph{Shared Facts} ($+5.20$\,pp), and \emph{Revisit Reasons} ($+5.06$\,pp).
Expansion is deepest for \emph{Track Evolution} (2.51 hops) and \emph{Revisit Reasons} (2.17), consistent with retrieving longer trajectories for evolution-focused questions rather than imposing a fixed depth.
\emph{Aligned Recommendations} is unchanged, while \emph{New Ideas} decreases by 3.27\,pp, showing that additional transition context does not uniformly benefit open-ended generation.
The 27B backbone improves across all seven categories; complete results and hop statistics are in Appendix~\ref{app:adaptive-hops}.

\section{Related Work}
\label{sec:related_work}

\paragraph{Agentic memory and memory management.}
Long-term agent memory has been studied as a way to extend LLM-based agents beyond stateless interaction.
Previous systems such as Generative Agents, MemoryBank, and MemGPT introduced persistent experience stores, reflection, hierarchical memory, and virtual context management for long-horizon interaction~\citep{generativeagents2023,memorybank2024,memgpt2023}.
Recent systems improve memory writing and consolidation through temporal graphs, scalable memory stores, self-organizing notes, staged consolidation, and recollection-style retrieval~\citep{zep2025,mem02025,amem2025,lightmem2026,rfmem2026}.
Long-term dialogue and personalization systems further use event memory, persona modeling, segment-level memory construction, and persistent user profiles~\citep{helloagain2025,pan2025secom,persistentprofiles2025}.
Other work studies modular, cognitive, or hierarchical memory organization for long-term agents and multi-agent systems~\citep{mirix2025,h-mem2026,o-mem2025,sun2025hierarchical,huang2025licomemory,li2025memos,zhang2025gmemory}.
A-Mem~\citep{amem2025} organizes memories as linked notes and updates their contents and connections as new experiences arrive.
\method{} instead stores explicit old-to-new state transitions and conditions subsequent graph expansion on the transitions already accepted along the retrieval path.

\paragraph{Memory retrieval.}
Retrieval-augmented generation retrieves chunks or passages that are semantically relevant to the query~\citep{rag2020}.
Temporal retrieval and time-aware memory organize evidence with timestamps, temporal knowledge graphs, or memory-augmented temporal reasoning modules~\citep{temporalsurvey2025,zep2025,SGMem2026,memotime2026}.
Graph-based retrieval further supports multi-hop evidence access through graph-augmented passage ranking, path retrieval, and graph walks~\citep{hipporag22025,pathrag2025,walkretrieve2025}.
These approaches improve access to relevant passages, facts, graph neighborhoods, or reasoning paths.
In evolving personal memory, however, the missing retrieval unit is often the preference update itself: which earlier state was superseded, what new state replaced it, why the change occurred, and when the new state became valid.
Recent memory systems also explore inference-time controllers for adaptive retrieval or reader feedback~\citep{remem2026,remindrag2025}, representing a different online-computation trade-off from the lightweight retrieval setting studied here.
\method{} instead moves preference-update interpretation to offline memory construction and retrieves stored transition records with embedding-based transition expansion at inference time.




\section{Conclusion}
\label{sec:conclusion}

We studied long-term personal memory under non-stationary user preferences. \method{} stores preference updates as transition records and retrieves them through factual anchoring followed by adaptive transition expansion. Across PersonaMem and PersonaBench, \method{} achieves the strongest overall scores under both Qwen backbones, and controlled analyses show that transition records and adaptive expansion add value beyond factual anchoring, chronological expansion, flat transition retrieval, and fixed-hop traversal. These results suggest that preference-update records are useful retrieval units for personal agents that must distinguish current user states from superseded ones as users change over time.


\section{Limitations}

\method{} relies on backbone language models for offline shift detection and final answer generation, so detector errors, hallucinations, or calibration failures may propagate into retrieval and responses.
Our evaluation focuses on text-based personalized-memory benchmarks and two open-source Qwen backbones; future work should test broader model families, multimodal histories, and longer deployment horizons.
As long-term personal memory may contain sensitive information and persistent memories can become stale or be manipulated, practical systems should support user inspection, correction, deletion, and access control~\citep{memorysecurity2026}.
A promising deployment direction is to store sensitive memory graphs locally or in user-controlled storage, exposing only the minimal retrieved evidence needed for each interaction.

\renewcommand{\acksection}{\section*{Acknowledgment}}
\begin{ack}
This research was supported through an Australian Research Council (ARC) Future Fellowship (FT210100097) and grant DP240101547.
\end{ack}

\bibliographystyle{plainnat}
\bibliography{references}

\appendix
\section{Dataset Details}
\label{app:datasets}

This work focuses on evolving personal preferences rather than general conversational memory.
While benchmarks such as LoCoMo~\citep{locomo2024}, LongMemEval~\citep{longmemeval2025}, and REALTALK~\citep{lee2025realtalk} evaluate broad long-term dialogue memory, we adopt PersonaMem and PersonaBench because they directly expose dynamic personalization and preference-oriented QA settings.

\subsection{PersonaMem}

PersonaMem~\citep{personamem2025} is a long-term personalized-memory benchmark constructed from multi-session dialogue histories in which user attributes, preferences, and contexts gradually evolve as a result of life events and changing situations.
Each user case consists of a chronological message stream paired with evaluation questions spanning seven categories that target different aspects of memory utilization and personalized generation; we adopt the released 32k-token and 128k-token settings.

\paragraph{User-case unit.}
In PersonaMem, the same persona can appear with different historical chat contexts. Since these histories are not necessarily identical, we use \texttt{shared\_context\_id} as the user-case unit. This ensures that each evaluation case is paired with its own personalized memory rather than merging distinct histories from the same persona name.

\paragraph{Data-quality filtering.}
We observed a form of context-independent answerability in which the question stem alone identifies the correct option without consulting the user history.
We use GPT-4o as a data-quality judge to detect and remove such items; the prompt is reproduced in Appendix~\ref{app:prompts}.
The numbers reported throughout the paper are computed on the same quality-filtered evaluation set for all methods.
Table~\ref{tab:personamem-distribution} compares the category composition before and after filtering.
The share of Track Evolution questions decreases from 23.60\% to 21.05\% at 32k and from 12.26\% to 10.18\% at 128k.

\begin{table}[h]
\centering
\caption{PersonaMem evaluation-set statistics after filtering. Token counts use \texttt{tiktoken} with \texttt{cl100k\_base}.}
\label{tab:personamem-overview}
\footnotesize
\setlength{\tabcolsep}{6pt}
\renewcommand{\arraystretch}{1.05}
\begin{tabular}{l r r}
\toprule
Statistic & 32k & 128k \\
\midrule
User cases & 37 & 55 \\
Questions & 475 & 2{,}161 \\
Avg.\ history tokens / user & 26{,}592.51 & 121{,}346.55 \\
Avg.\ messages / user & 174.65 & 761.91 \\
\bottomrule
\end{tabular}
\end{table}

\begin{table}[h]
\centering
\caption{Evaluation aspects in PersonaMem.}
\label{tab:personamem-aspects}
\footnotesize
\setlength{\tabcolsep}{4pt}
\renewcommand{\arraystretch}{1.10}
\begin{tabular}{l p{0.78\linewidth}}
\toprule
Aspect & Evaluates \\
\midrule
Revisit Reasons  & Recall the reason or event that caused a user preference to change. \\
Track Evolution  & Recover how a user preference changes across time rather than only the latest state. \\
Latest Prefs     & Recognize the most recent preference expressed in the user history. \\
Aligned Recs     & Recommend items or activities aligned with the user's current preferences. \\
New Scenario     & Apply user preferences to a new situation not directly seen in the history. \\
Shared Facts     & Recall static events, activities, or interests the user previously shared. \\
New Ideas        & Suggest new items or activities not already mentioned when the user asks for novelty. \\
\bottomrule
\end{tabular}
\end{table}

\begin{table}[h]
\centering
\caption{PersonaMem query-type distribution before and after filtering. Each cell reports the count and its percentage within the corresponding split.}
\label{tab:personamem-distribution}
\footnotesize
\setlength{\tabcolsep}{4pt}
\renewcommand{\arraystretch}{1.05}
\begin{tabular}{l r r r r}
\toprule
& \multicolumn{2}{c}{32k} & \multicolumn{2}{c}{128k} \\
\cmidrule(lr){2-3} \cmidrule(lr){4-5}
Aspect & Before & After & Before & After \\
\midrule
Revisit Reasons & 99 (16.81\%)  & 54 (11.37\%)  & 237 (9.69\%)   & 158 (7.31\%) \\
Track Evolution & 139 (23.60\%) & 100 (21.05\%) & 300 (12.26\%)  & 220 (10.18\%) \\
Latest Prefs    & 17 (2.89\%)   & 17 (3.58\%)   & 772 (31.56\%)  & 689 (31.88\%) \\
Aligned Recs    & 55 (9.34\%)   & 54 (11.37\%)  & 320 (13.08\%)  & 310 (14.35\%) \\
New Scenario    & 57 (9.68\%)   & 49 (10.32\%)  & 195 (7.97\%)   & 171 (7.91\%) \\
Shared Facts    & 129 (21.90\%) & 113 (23.79\%) & 155 (6.34\%)   & 154 (7.13\%) \\
New Ideas       & 93 (15.79\%)  & 88 (18.53\%)  & 467 (19.09\%)  & 459 (21.24\%) \\
\midrule
Total           & 589 (100\%)   & 475 (100\%)   & 2{,}446 (100\%) & 2{,}161 (100\%) \\
\bottomrule
\end{tabular}
\end{table}

\subsection{PersonaBench}

PersonaBench~\citep{personabench} evaluates personalized QA over simulated private user data. We use the official six-user setting from the release, comprising 263 QA items. Each user's private data consists of short multi-turn memory segments (87.83 per user on average) rather than a single raw chat transcript.
It evaluates four aspects:
\emph{Basic information} that covers stable personal attributes and basic facts,
\emph{Preference easy} that targets direct preference evidence,
\emph{Preference hard} requiring harder preference reasoning beyond direct lookup, and \emph{Social} which covers social attributes and relationships derived from private user data.
We report lexical F1 following the official evaluation protocol.
We reuse the construction and retrieval hyperparameters selected on PersonaMem-32k and apply a uniform 1,000-token retrieved-context budget to all methods.

\subsection{Evaluations}
\label{app:evaluation}

\textbf{PersonaMem}~\citep{personamem2025} evaluates QA on four-option questions.
For each question type $c$ with $n_c$ questions answered at accuracy $a_c$ (in \%), the global overall accuracy is the weighted average across all types:
\[
\text{Overall} = \frac{\displaystyle\sum_{c} a_c \times n_c}{\displaystyle\sum_{c} n_c}.
\]


End-to-end QA performance on \textbf{PersonaBench}~\citep{personabench} is measured with lexical F1 using the benchmark's official evaluation protocol.
Note that Recall@$k$ is not used as the metric because the retrieval unit is not comparable across memory systems: \method{} retrieves atomic transition records and normal facts, whereas chunk-based baselines retrieve much larger segments.

After applying the official answer preprocessing, let $G_i$ and $H_i$ denote the sets of unique gold and predicted tokens for example $i$.
We compute
\begin{equation}
\mathrm{Prec}_i =
\begin{cases}
\frac{|G_i \cap H_i|}{|H_i|}, & |H_i| > 0,\\
0, & |H_i| = 0,
\end{cases}
\qquad
\mathrm{Rec}_i =
\begin{cases}
\frac{|G_i \cap H_i|}{|G_i|}, & |G_i| > 0,\\
0, & |G_i| = 0.
\end{cases}
\end{equation}
The per-example F1 is
\begin{equation}
\mathrm{F1}_i =
\begin{cases}
\frac{2\,\mathrm{Prec}_i\,\mathrm{Rec}_i}{\mathrm{Prec}_i+\mathrm{Rec}_i}, & \mathrm{Prec}_i+\mathrm{Rec}_i > 0,\\
0, & \mathrm{Prec}_i+\mathrm{Rec}_i = 0.
\end{cases}
\end{equation}

\section{Implementation Details}
\label{app:impl-details}

\subsection{Implementation Setup}

\paragraph{Models.}
We use two backbones, Qwen3-8B~\citep{qwen3_2025} and Qwen3.5-27B~\citep{qwen35_2026}. Each backbone is used for memory graph construction, shift detection ($\mathcal{D}_{\mathrm{LLM}}$), and inference-time QA.

\paragraph{Retriever.}
We use BGE-M3~\citep{M3embedding2024} as the dense retriever for all embedding-based computations in both the construction and retrieval stages.

\paragraph{Graph construction engine.}
We use a modified Graphiti\footnote{\url{https://github.com/getzep/graphiti}} implementation (Apache-2.0) as the underlying graph construction engine.

\paragraph{Hardware.}
All experiments are conducted on a single NVIDIA A100 GPU.

\paragraph{Construction cost.}
Table~\ref{tab:construction-cost} reports the average per-user token cost for offline memory construction, measured on PersonaMem 32k with Qwen3.5-27B.
\method{} is substantially cheaper than \textsc{A-Mem} (722k vs.\ 950k tokens per user, a $1.31\times$ reduction) while achieving higher overall accuracy; it is more expensive than \textsc{Mem0} but recovers a clear accuracy advantage through preference-trajectory retrieval.

\begin{table}[h]
\centering
\caption{Average per-user token cost for offline memory construction (PersonaMem 32k, Qwen3.5-27B).}
\label{tab:construction-cost}
\footnotesize
\setlength{\tabcolsep}{6pt}
\renewcommand{\arraystretch}{1.05}
\begin{tabular}{l r r r}
\toprule
Method & Input & Output & Total \\
\midrule
\textsc{Mem0}     & 434,977 & 121,250 & 556,227 \\
\textsc{A-Mem}    & 867,453 &  82,513 & 949,966 \\
\method{}~(ours)  & 620,976 & 101,201 & 722,177 \\
\bottomrule
\end{tabular}
\end{table}

\subsection{Memory Baselines}
\label{app:baselines}

\paragraph{RF-Mem~\citep{rfmem2026}.}
RF-Mem models retrieval through a familiarity--recollection mechanism that first performs lightweight top-$k$ retrieval and then selectively expands evidence through a slower recollection path when uncertainty remains.
It is a strong retrieval-oriented baseline because it adapts evidence collection at inference time without explicitly constructing or traversing preference-transition edges.

\paragraph{Zep~\citep{zep2025}.}
Zep builds agent memory on top of a temporal knowledge graph, converting conversations into entities, relationships, and temporally scoped facts via its Graphiti engine.
Outdated facts are invalidated rather than deleted, so provenance is preserved.
We reproduce Zep from its open-source release rather than its commercial API.

\paragraph{A-Mem~\citep{amem2025}.}
A-Mem introduces an agentic memory system inspired by the Zettelkasten method, converting each interaction into a structured note with contextual descriptions and keywords, then dynamically linking it to related historical memories.
Unlike flat vector stores, A-Mem emphasizes note-level self-organization and memory updating, making it a meaningful comparison for preference-tracking tasks.

\paragraph{Mem0~\citep{mem02025}.}
Mem0 is a scalable long-term memory architecture that dynamically extracts, consolidates, updates, and retrieves salient information from ongoing conversations.
We use the graph-enhanced variant, which additionally represents relational structures among memory items, and reproduce it from the open-source release rather than the commercial API.

\paragraph{LightMem~\citep{lightmem2026}.}
LightMem is an efficiency-oriented memory framework using a three-stage pipeline inspired by the Atkinson--Shiffrin model: sensory memory filters raw interactions, short-term memory groups and summarizes by topic, and long-term memory performs offline consolidation.
This design reduces online overhead and is a strong baseline for understanding whether structured compression alone suffices in long-context settings.

\subsection{Intra- and inter-topic transitions}

Preference updates are not always expressed within a single stable topic boundary. 
Such updates may occur within a fine-grained topic graph, such as revising a fitness routine, or across related topic graphs, such as a change in social life altering what outdoor activities the user now prefers.
We therefore label grounded transition edges by the topic assignment of their endpoints, which helps separate local refinements from updates whose consequences cross fine-grained topic graphs.

\paragraph{Intra-topic transitions.}
An intra-topic transition captures a preference update whose old and new states are grounded within the same topic graph.
For example, a user may initially prefer short morning workouts but later state that they now prefer longer evening strength sessions after changing their work schedule.
Both states belong to a \texttt{fitness-routine} topic, but the transition record is still necessary because the later state revises how future fitness-related recommendations should be interpreted.
Thus, intra-topic transitions primarily model local refinements, replacements, or reversals within a stable topical context.

\paragraph{Inter-topic transitions.}
Some preference updates are expressed in one topical context but change how memories in another topic should be interpreted.
For example, a user may initially prefer quick solo workouts because of a busy schedule, but later prefer longer outdoor activities after joining a weekend hiking group.
The old state may be grounded under a \texttt{fitness-routine} topic, whereas the new state may be grounded under a \texttt{social-outdoor-activities} topic.
Both endpoints are produced by the same shift detector; the edge is labeled inter-topic only because the grounded endpoints belong to different topic graphs.
Thus, inter-topic transitions model preference updates whose consequences cross fine-grained topic boundaries.

Figure~\ref{fig:ablation} evaluates the contribution of these two edge types separately, showing that intra-topic and inter-topic transitions provide complementary retrieval signals.

\subsection{Transition Records}
\label{app:transition-records}
The detector outputs shift events, which are stored as transition records after grounding. 
Table~\ref{tab:transition-record-examples} shows examples of the stored fields.

\paragraph{Grounding threshold.}
We set $\tau_g=0.50$ and discard a detected update if either endpoint similarity falls below this threshold.
This lightweight check requires both extracted states to have sufficiently similar anchors in the existing graph.
It can reject extraction hallucinations that lack such anchors without an additional LLM call.

\paragraph{Illustrative grounding example.}
The message ``I prefer quick solo workouts'' introduces entity nodes such as \texttt{User} and \texttt{solo workouts}, with a factual edge \texttt{User} $\xrightarrow{\mathrm{PREFERS}}$ \texttt{solo workouts} in a fitness topic graph.
A later message expressing a preference for social outdoor activities is also ingested as factual memory.
The shift detector produces a record with $s^{-}=$ ``quick solo workouts'' and $s^{+}=$ ``social outdoor activities''.
Equation~\ref{eq:grounding} maps these states to the existing \texttt{solo workouts} and \texttt{social outdoor activities} nodes.
If both similarities meet $\tau_g$, we store the record and its directed edge $v^{-}\!\to v^{+}$; the edge is inter-topic when the endpoints belong to different topic graphs.
The state descriptions remain in the record rather than becoming new nodes.

\begin{table}[h]
\centering
\caption{Examples of transition records produced by the shift detector.}
\label{tab:transition-record-examples}
\footnotesize
\setlength{\tabcolsep}{3pt}
\renewcommand{\arraystretch}{1.08}
\begin{tabular}{@{}p{0.13\linewidth}p{0.21\linewidth}p{0.21\linewidth}p{0.26\linewidth}p{\dimexpr0.19\linewidth-8\tabcolsep\relax}@{}}
\toprule
Type & Old state & New state & Update rationale & Valid time \\
\midrule
Music preference shift &
Engaging with the quirky and humorous side of music &
Focusing on emotional and reflective aspects of music &
The user moved from lighthearted storytelling to a more introspective music experience. &
2015-03-22 \\
Learning approach shift &
Learning through formal, high-density sources &
Preferring accessible, informal, discussion-friendly sources &
Jargon-heavy material caused confusion, while informal mentoring worked better. &
2018-09-14 \\
\bottomrule
\end{tabular}
\end{table}

\subsection{Algorithm Workflow}
\label{app:algorithms}

Algorithms~\ref{alg:build} and~\ref{alg:infer} restate the construction and retrieval stages of \method{} in pseudocode, making the order of operations and all threshold dependencies explicit.

\begin{algorithm}[t]
\caption{Topic-Structured Preference Graph Construction.}
\label{alg:build}
\begin{algorithmic}[1]
\Require user history $m_1,\dots,m_T$;\; thresholds $\tau_{\mathrm{macro}},\,\tau_{\mathrm{micro}},\,\kappa$;\; grounding threshold $\tau_g=0.50$;\; maturity bound $n_{\mathrm{mat}}$;\; historical-evidence budget $K_{\mathrm{hist}}$
\Ensure topic-structured user memory graph $\usergraph$
\State initialize topic set $\{T_i\}$ and candidate-update buffer $\evobuf_u \gets \emptyset$
\For{each incoming message $m_t$}
    \State compute $S_{\mathrm{macro}}(m_t,T_i)$ for all existing topic graphs $T_i$
    \State $M_t \gets \max_i S_{\mathrm{macro}}(m_t,T_i)$;\quad $i^* \gets \arg\max_i S_{\mathrm{macro}}(m_t,T_i)$
    \If{$M_t < \tau_{\mathrm{macro}}$}
        \State $r_t \gets \dbirth$;\quad create a new topic graph and ingest $m_t$
    \Else
        \State $\mu_t \gets S_{\mathrm{micro}}(m_t,T_{i^*})$;\quad $z_t \gets \mathrm{mature}(T_{i^*}; n_{\mathrm{mat}})$
        \If{$\neg\, z_t$ \textbf{or} $\mu_t \ge \tau_{\mathrm{micro}}$}
            \State $r_t \gets \ddeepen$;\quad ingest $m_t$ into $T_{i^*}$
        \Else
            \State $r_t \gets \devolve$;\quad ingest $m_t$ into $T_{i^*}$
            \State $\evobuf_u \gets \evobuf_u \cup \{m_t\}$
            \If{$|\evobuf_u| \ge \kappa$}
                \State $\mathcal{H}(\evobuf_u) \gets \operatorname*{arg\,topK}_{f \in \mathcal{F}_u,\;K_{\mathrm{hist}}} s_{\mathrm{hist}}(\evobuf_u,f)$
                \Comment{retrieve compact historical evidence from the user graph}
                \State $\mathcal{Z} \gets \mathcal{D}_{\mathrm{LLM}}\!\left(\evobuf_u,\mathcal{H}(\evobuf_u)\right)$
                \For{each detected update $z=(s^{-},s^{+},r,g,t_{\mathrm{valid}})\in\mathcal{Z}$}
                    \State map $s^{-},s^{+}$ to $v^{-},v^{+}\in V_u$ using Equation~\ref{eq:grounding}
                    \If{both endpoint similarities are at least $\tau_g$}
                        \State set $y_e=\mathrm{intra}$ if endpoints share a topic, otherwise $y_e=\mathrm{inter}$
                        \State store $e_z=(v^{-},v^{+},r,y_e,g,t_{\mathrm{valid}})$ and its record $z$
                    \Else
                        \State discard $z$
                    \EndIf
                \EndFor
                \State $\evobuf_u \gets \emptyset$
            \EndIf
        \EndIf
    \EndIf
    \State periodically refresh affected topic-summary embeddings $s_i$
\EndFor
\State \Return $\usergraph$
\end{algorithmic}
\end{algorithm}

Algorithm~\ref{alg:build} processes user history in a single pass. Every message is routed and ingested; messages routed as \textsc{Evolve} are added to a candidate-update buffer $\mathcal{B}$ that triggers detection once $\kappa$ candidates accumulate. For each triggered buffer, \method{} retrieves a compact historical-evidence set $\mathcal{H}(\mathcal{B})$ from the user memory graph before calling the shift detector. This historical-evidence retrieval is capped at $K_{\mathrm{hist}}=10$ normal facts, so the detector compares the candidate buffer against a small evidence set rather than the full user history.
The three pseudocode labels correspond to the routing outcomes of Equation~\ref{eq:routing}: \dbirth{} instantiates a new topic when the macro gate fails; \ddeepen{} continues an existing topic when both gates pass; \devolve{} flags an evolution candidate when only the macro gate passes.
The shift detector prompt used by $\mathcal{D}_{\mathrm{LLM}}$ is shown in Appendix~\ref{app:prompts}.

\paragraph{Maturity bound.}
We enable the \textsc{Evolve} branch only after the matched topic graph contains at least $n_{\mathrm{mat}}$ ingested messages.
This safeguard prevents sparse early topic graphs from producing unstable shift candidates; it does not introduce a separate detector or transition type.
In the early phase of a topic, the entity set $K(T_{i^*})$ is small, so the entity-overlap score $\mu_t = S_{\mathrm{micro}}(m_t, T_{i^*})$ is naturally low simply because the topic has accumulated few entities, not because the new message conflicts with the existing state.
Without the maturity check, such low-overlap messages would immediately route to \devolve{} and produce spurious candidate updates against an essentially empty topic.
Before the topic matures, all topically related messages are therefore routed to \ddeepen{}; once it matures, the macro--micro routing in Equation~\ref{eq:routing} applies as written.
We adopt $n_{\mathrm{mat}}{=}15$ and refresh topic summaries after every 10 topic-local deepening messages, counting each message as one update unit.
These values are treated as supporting implementation settings rather than as primary experimental axes; sensitivity analyses for the main construction and retrieval hyperparameters are reported in Appendix~\ref{app:build-hypara} and Appendix~\ref{app:infer-hypara}.

\begin{algorithm}[t]
\caption{Adaptive Preference Retrieval.}
\label{alg:infer}
\begin{algorithmic}[1]
\Require query $q$;\; graph $\usergraph$;\; factual-anchor budget $K_A$;\; continuation threshold $\tau_c$;\; drop tolerance $\delta_c$;\; retention $\gamma$;\; beam size $B$;\; max hops $H$
\Ensure evidence context $\ctxset(q)$
\State $\mathcal{A}(q) \gets \operatorname*{arg\,topK}_{f \in \mathcal{F}_{\mathrm{normal}},\;K_A} s(q,f)$
    \Comment{retrieve factual anchors}
\State $\mathcal{N}_0(q) \gets$ transition-edge candidates adjacent to the endpoints of factual anchors in $\mathcal{A}(q)$
\For{each $e \in \mathcal{N}_0(q)$}
    \State $\sigma_0(e) \gets \cos\!\left(\embed(q),\embed(e)\right)$
\EndFor
\State $\widetilde{\mathcal{N}}_0(q) \gets \{e \in \mathcal{N}_0(q): \sigma_0(e) \ge \tau_c\}$
\State initialize $\mathcal{P}$ from the top-$B$ candidates in $\widetilde{\mathcal{N}}_0(q)$
\If{$\mathcal{P} = \emptyset$}
    \State \Return $\mathrm{Assemble}\!\left(\mathcal{A}(q),\emptyset\right)$
        \Comment{no relevant transition found}
\EndIf
\For{each path $p \in \mathcal{P}$}
    \State $C^{(p)} \gets \embed(q)$;\quad $\sigma_{\mathrm{last}}^{(p)} \gets \sigma_0(e_1^{(p)})$
    \State $C^{(p)} \gets \gamma C^{(p)} + (1-\gamma)\embed(e_1^{(p)})$
\EndFor
\State $\mathcal{P}_{\mathrm{done}} \gets \emptyset$
\For{$t = 2,\dots,H$}
    \State $\mathcal{P}' \gets \emptyset$
    \For{each active path $p \in \mathcal{P}$}
        \State $\mathcal{N}(p) \gets$ transition-edge candidates adjacent to the current endpoint of $p$
        \For{each candidate next-hop edge $e \in \mathcal{N}(p)$}
            \State $\sigma(e\mid p) \gets \cos\!\left(C^{(p)},\embed(e)\right)$
            \If{$\sigma(e\mid p) \ge \tau_c\;\wedge\;\sigma(e\mid p)-\sigma_{\mathrm{last}}^{(p)} \ge -\delta_c$}
                \State extend $p$ to $p'$ with $e$
                \State $C^{(p')} \gets \gamma C^{(p)} + (1-\gamma)\embed(e)$
                \State $\sigma_{\mathrm{last}}^{(p')} \gets \sigma(e\mid p)$
                \State add $p'$ to $\mathcal{P}'$
            \EndIf
        \EndFor
        \If{no candidate extension of $p$ is accepted}
            \State $\mathcal{P}_{\mathrm{done}} \gets \mathcal{P}_{\mathrm{done}} \cup \{p\}$
        \EndIf
    \EndFor
    \If{$\mathcal{P}' = \emptyset$}
        \State \textbf{break}
    \EndIf
    \State $\mathcal{P} \gets \mathrm{TopB}(\mathcal{P}')$
\EndFor
\State $\mathcal{E}_{\mathrm{trans}}(q) \gets$ unique transition records along paths in $\mathcal{P} \cup \mathcal{P}_{\mathrm{done}}$
\State \Return $\mathrm{Assemble}\!\left(\mathcal{A}(q),\mathcal{E}_{\mathrm{trans}}(q)\right)$
\end{algorithmic}
\end{algorithm}

Algorithm~\ref{alg:infer} implements the factual anchoring followed by adaptive transition expansion described in Section~\ref{sec:retrieval}.
After retrieving factual anchors, all transition-edge candidates adjacent to the anchor entities are scored directly against the query embedding; the top-$B$ candidates seed the beam.
If no candidate scores above the continuation threshold $\tau_c$, retrieval terminates at the factual anchors.
Otherwise, each accepted path extends under a shared continuation rule $(\tau_c, \delta_c)$ and trajectory state updated by the retention factor $\gamma$; beam size $B$ and hop cap $H$ act as second-line safeguards (see Appendix~\ref{app:infer-hypara}).
At each hop, $\mathrm{TopB}$ retains up to $B$ paths with the highest latest-step scores $\sigma_{\mathrm{last}}^{(p)}$.

\subsection{Prompt Templates}
\label{app:prompts}

We include the prompt templates used for transition detection, transition-record quality analysis, final QA on PersonaMem and PersonaBench, and PersonaMem data-quality filtering.
Prompts are shown verbatim; line breaks are preserved for readability.

\paragraph{Shift detector prompt.}
$\mathcal{D}_{\mathrm{LLM}}$ (Equation~\ref{eq:detector}) is prompted as follows.

\begin{tcolorbox}[
  colback=gray!4, colframe=black!60,
  title={\footnotesize\bfseries Shift Detector (Memory Detective)},
  fonttitle=\footnotesize\bfseries,
  coltitle=black,
  arc=2mm, boxrule=0.6pt,
  left=4pt, right=4pt, top=3pt, bottom=3pt,
  breakable
]
\scriptsize
You are a specialized Memory Detective. Your goal is to analyze a timeline of user
activities and identify \textbf{``Shift Events''} — distinct points where the user's
preferences, habits, or state \textbf{changed} from one to another.

\medskip
\textbf{Input Context (Chronological Facts):}

\{sorted\_facts\}

\medskip
\textbf{Instructions:}
\begin{enumerate}
  \item Read the timeline to detect contradictions or explicit updates over time.
  \item Ignore simple refinements. Detect actual state changes.
  \item For each shift, extract: event\_type, old\_state, new\_state, trigger, valid\_at (YYYY-MM-DD).
  \item If NO significant shifts found, return an empty list.
\end{enumerate}

\textbf{Output Format (strict JSON):}

\verb|{"shifts": [{"event_type": "...", "old_state": "...",|

\verb|  "new_state": "...", "trigger": "...", "valid_at": "YYYY-MM-DD"}]}|
\end{tcolorbox}

\paragraph{LLM-as-judge prompt for transition-record quality analysis.}
The judge prompt used in Appendix~\ref{app:build-hypara} is shown below.

\begin{tcolorbox}[
  colback=gray!4, colframe=black!60,
  title={\footnotesize\bfseries Transition-Record Judge (GPT-4o)},
  fonttitle=\footnotesize\bfseries,
  coltitle=black,
  arc=2mm, boxrule=0.6pt,
  left=4pt, right=4pt, top=3pt, bottom=3pt,
  breakable
]
\scriptsize
\textbf{System:} You are evaluating shift records in a personalized lifelong memory system.
A shift record claims a user's preference or behavior has CHANGED over time.

Rate VALID if ALL of the following hold:
\begin{enumerate}
  \item The old state and new state describe meaningfully DIFFERENT preferences or behaviors.
  \item The trigger (if present) plausibly explains the change.
  \item The new state is specific enough to support future personalized retrieval.
\end{enumerate}
Rate INVALID if ANY of the following hold:
old/new states are essentially the same (trivial rewording);
change is cessation only with no new preference;
trigger does not plausibly cause the change;
states are incoherent or self-contradictory;
both states are too vague to be useful for retrieval.

Reply with exactly: Line 1: VALID or INVALID. Line 2: one-sentence reason ($\leq$30 words).

\medskip
\textbf{Template:} Shift type: \{kind\} $\mid$ Fact: \{fact\} $\mid$ Trigger: \{trigger\} $\mid$
Source: \{src\_name\} $\mid$ Target: \{tgt\_name\} $\mid$ VALID or INVALID?
\end{tcolorbox}

\paragraph{Final QA prompt for PersonaMem.}
At inference time, the QA backbone is asked to choose one of the four candidate options using the retrieved memory context.

\begin{tcolorbox}[
  colback=gray!4, colframe=black!60,
  title={\footnotesize\bfseries Final QA Prompt (PersonaMem)},
  fonttitle=\footnotesize\bfseries,
  coltitle=black,
  arc=2mm, boxrule=0.6pt,
  left=4pt, right=4pt, top=3pt, bottom=3pt,
  breakable
]
\scriptsize
\textbf{Relevant historical information:}\\
\{context\_text\}

\medskip
\textbf{Question:}\\
\{question\}

\medskip
\textbf{Candidate options:}\\
(a) \{option\_a\}\\
(b) \{option\_b\}\\
(c) \{option\_c\}\\
(d) \{option\_d\}

\medskip
Find the most appropriate model response and only give your final answer as
\verb|<final_answer>(a)</final_answer>|, \verb|<final_answer>(b)</final_answer>|,
\verb|<final_answer>(c)</final_answer>|, or \verb|<final_answer>(d)</final_answer>|.
Do not provide any explanation.
\end{tcolorbox}

\paragraph{Final QA prompt for PersonaBench.}
PersonaBench answers are short spans rather than multiple-choice options, so the prompt asks for a minimal answer drawn from the retrieved history.

\begin{tcolorbox}[
  colback=gray!4, colframe=black!60,
  title={\footnotesize\bfseries Final QA Prompt (PersonaBench)},
  fonttitle=\footnotesize\bfseries,
  coltitle=black,
  arc=2mm, boxrule=0.6pt,
  left=4pt, right=4pt, top=3pt, bottom=3pt,
  breakable
]
\scriptsize
\textbf{Relevant historical information:}\\
\{context\_text\}

\medskip
\textbf{Question:} \{question\}

\medskip
Answer using only a short answer span copied or normalized from the history when possible.
Do not explain. Do not use a full sentence unless the answer itself is naturally a sentence.
Do not add extra details, hedging, or commentary. Output only the final answer.
\end{tcolorbox}

\paragraph{PersonaMem quality prompt.}
We use GPT-4o as a data-quality judge to detect PersonaMem items where the question stem alone reveals the correct option without consulting the user history (Appendix~\ref{app:datasets}).

\begin{tcolorbox}[
  colback=gray!4, colframe=black!60,
  title={\footnotesize\bfseries PersonaMem Quality Prompt},
  fonttitle=\footnotesize\bfseries,
  coltitle=black,
  arc=2mm, boxrule=0.6pt,
  left=4pt, right=4pt, top=3pt, bottom=3pt,
  breakable
]
\scriptsize
You are an expert Data Quality Auditor.
Your task is to detect ``Data Leakage'' in multiple-choice questions.

\medskip
\textbf{Definitions:}

\textbf{LEAKED (True):} The Question Stem explicitly contains key information (e.g., specific user preferences, past events, names) that directly reveals the Correct Answer. The user does NOT need to recall any past memory to answer this.

\textbf{CLEAN (False):} The Question Stem provides context but does NOT give away the answer. The user MUST recall external memory/history to be sure.
\textit{CRITICAL:} If the answer can be guessed purely by random chance (25\%) or general common sense, it is NOT a leak. Mark it as CLEAN. Only mark LEAKED if the text itself provides the evidence.

\medskip
\textbf{Question:} \{question\}

\medskip
\textbf{Output Format (strict JSON):}

\verb|{"reasoning": "Step-by-step analysis. Does the stem mention X?",|

\verb| "is_leaked": true or false}|
\end{tcolorbox}

\section{Additional Experimental Results}
\label{app:additional-results}

\subsection{Full-Context Baseline}
\label{app:fulltext}

\textsc{FullText} is a prompt-only method that prepends the user's complete conversational history before each query, giving the model direct read access to all past interactions without retrieval or memory management.
Table~\ref{tab:fulltext} reports its results across all four backbone\,$\times$\,context configurations.

\begin{table}[h]
\centering
\caption{\textsc{FullText} results across all configurations. The 128k\,+\,Qwen3-8B entry is unavailable because the full history exceeds the model's context limit.}
\label{tab:fulltext}
\footnotesize
\setlength{\tabcolsep}{4pt}
\renewcommand{\arraystretch}{1.05}
\begin{tabular}{ll c c c c c c c c}
\toprule
Context & Model & \makecell{Revisit\\Reasons} & \makecell{Track\\Evo} & \makecell{Latest\\Prefs} & \makecell{Aligned\\Recs} & \makecell{New\\Scenario} & \makecell{Shared\\Facts} & \makecell{New\\Ideas} & Overall \\
\midrule
\multirow{2}{*}{32k}  & Qwen3-8B      & 62.96 & 64.00 & 47.06 & 42.59 & 53.06 & 46.02 & 10.23 & 45.47 \\
                      & Qwen3.5-27B   & 90.74 & 77.00 & 58.82 & 61.11 & 77.55 & 69.03 & 27.27 & 65.05 \\
\midrule
\multirow{2}{*}{128k} & Qwen3-8B      & \multicolumn{8}{c}{--- \emph{(exceeds backbone context limit)}} \\
                      & Qwen3.5-27B   & 89.24 & 77.73 & 60.81 & 52.58 & 76.61 & 71.43 & 33.33 & 59.60 \\
\bottomrule
\end{tabular}
\end{table}

\textsc{FullText} with Qwen3.5-27B achieves the strongest results on \emph{Revisit Reasons} and \emph{Track Evolution}, where direct access to the complete record eliminates retrieval noise.
However, every query requires forwarding the full history through the model; token consumption scales linearly with history length, and the 128k\,+\,Qwen3-8B configuration is infeasible because the model's 32k context limit is exceeded.
As user histories grow longer, this hard limit becomes increasingly restrictive.

\subsection{By-Task Results for \method{}}
\label{app:by-topic}

Tables~\ref{tab:bytopic-32k} and~\ref{tab:bytopic-128k} report per-task accuracy of \method{} across both backbones.
Topic abbreviations: \emph{Food} (foodRecommendation), \emph{Home} (homeDecoration), \emph{Medical} (medicalConsultation), \emph{Travel} (travelPlanning), \emph{Legal} (legalConsultation), \emph{Financial} (financialConsultation), \emph{Dating} (datingConsultation), \emph{Music} (musicRecommendation), \emph{Book} (bookRecommendation), \emph{Study} (studyConsultation), \emph{Movie} (movieRecommendation), \emph{Family} (familyRelations), \emph{Therapy} (therapy), \emph{Shopping} (onlineShopping), \emph{Sports} (sportsRecommendation).
The 128k benchmark adds \emph{Shopping} and \emph{Sports} not present in the 32k partition.
Gains are not concentrated in a single domain; the magnitude of improvement varies across topics, consistent with differences in preference-update density and temporal disambiguation difficulty.

\begin{table}[h]
\centering
\caption{Per-task accuracy of \method{} in the 32k context regime. $n$ = evaluation queries per topic.}
\label{tab:bytopic-32k}
\footnotesize
\setlength{\tabcolsep}{6pt}
\begin{tabular}{l r r r}
\toprule
Topic & $n$ & Qwen3-8B & Qwen3.5-27B \\
\midrule
Food      &  2 & 100.00 & 100.00 \\
Home      &  2 & 100.00 & 100.00 \\
Medical   & 14 & \phantom{0}92.86 & \phantom{0}92.86 \\
Legal     & 28 & \phantom{0}64.29 & \phantom{0}75.00 \\
Travel    & 60 & \phantom{0}60.00 & \phantom{0}70.00 \\
Study     & 29 & \phantom{0}58.62 & \phantom{0}62.07 \\
Dating    & 67 & \phantom{0}58.21 & \phantom{0}70.15 \\
Movie     & 84 & \phantom{0}57.14 & \phantom{0}71.43 \\
Book      & 58 & \phantom{0}56.90 & \phantom{0}67.24 \\
Music     & 45 & \phantom{0}55.56 & \phantom{0}60.00 \\
Therapy   &  2 & \phantom{0}50.00 & \phantom{0}50.00 \\
Financial & 60 & \phantom{0}45.00 & \phantom{0}73.33 \\
Family    & 24 & \phantom{0}37.50 & \phantom{0}54.17 \\
\midrule
\textit{Overall} & 475 & \textit{56.84} & \textit{69.26} \\
\bottomrule
\end{tabular}
\end{table}

\begin{table}[h]
\centering
\caption{Per-task accuracy of \method{} in the 128k context regime. $n$ = evaluation queries per topic.}
\label{tab:bytopic-128k}
\footnotesize
\setlength{\tabcolsep}{6pt}
\begin{tabular}{l r r r}
\toprule
Topic & $n$ & Qwen3-8B & Qwen3.5-27B \\
\midrule
Home      & 111 & 58.56 & 63.96 \\
Study     & 131 & 58.02 & 74.81 \\
Family    & 117 & 55.56 & 70.94 \\
Travel    & 126 & 53.97 & 65.87 \\
Music     &  38 & 52.63 & 73.68 \\
Medical   & 160 & 51.25 & 73.12 \\
Shopping  & 164 & 51.22 & 56.10 \\
Movie     &  75 & 49.33 & 70.67 \\
Book      & 125 & 47.20 & 72.00 \\
Dating    & 170 & 46.47 & 63.53 \\
Financial & 175 & 45.71 & 66.29 \\
Legal     & 256 & 45.70 & 60.94 \\
Food      & 198 & 41.41 & 60.61 \\
Sports    & 114 & 41.23 & 56.14 \\
Therapy   & 201 & 38.31 & 59.20 \\
\midrule
\textit{Overall} & 2161 & \textit{48.03} & \textit{64.69} \\
\bottomrule
\end{tabular}
\end{table}

\subsection{Various Retrieval Strategies}
\label{app:retrieval-strategies}

We include two inference-only controls to separate the effect of transition-aware retrieval from simpler context-expansion alternatives. Both controls reuse the same constructed memory and final QA prompt as \method{}, but replace adaptive transition expansion with a simpler retrieval strategy.

The three expansion-control variants compare as follows:
\begin{itemize}
  \item \textsc{ChronoExpand}: factual anchors plus chronological normal-fact neighbors.
  \item \textsc{FlatTrans}: factual anchors plus flat retrieved transition records.
  \item \method{}: factual anchors plus graph-grounded adaptive transition expansion.
\end{itemize}

\paragraph{ChronoExpand.}
\label{app:chrono}
\textsc{ChronoExpand} starts from the same factual anchors as \method{}.
For each anchor fact, it locates the anchor in the same topic's fact timeline (sorted by \texttt{valid\_at}/\texttt{created\_at}), retrieves the $h$ nearest normal facts on each side, and merges the result into the final context under the same token budget.
It does not use transition records and does not follow transition endpoints.
This tests whether nearby temporal context alone can explain the gains of \method{}.

For example, if an anchor fact states that the user recently joined a hiking group, \textsc{ChronoExpand} may also retrieve the immediately preceding and following normal facts in the same outdoor-activity timeline. It does not know whether those neighboring facts describe a preference update or merely adjacent context.

We select $h$ on a 20-user diagnostic subset (257 queries) using the same procedure as inference hyperparameter search (Appendix~\ref{app:infer-hypara}).
Table~\ref{tab:chrono-h} reports overall accuracy for $h \in \{1, 2, 3\}$.

\begin{table}[h]
\centering
\caption{\textsc{ChronoExpand} overall accuracy on the diagnostic subset (20 users, 257 queries) for different chronological window sizes.}
\label{tab:chrono-h}
\footnotesize
\setlength{\tabcolsep}{8pt}
\renewcommand{\arraystretch}{1.05}
\begin{tabular}{c r}
\toprule
Window $h$ & Overall Acc.\ (\%) \\
\midrule
1 & 55.64 \\
\textbf{2} & \textbf{56.03} \\
3 & 53.78 \\
\bottomrule
\end{tabular}
\end{table}

$h{=}2$ yields the highest accuracy and is used for all \textsc{ChronoExpand} comparisons in Section~\ref{sec:hops}. Wider windows can add temporally adjacent but answer-irrelevant normal facts, diluting the answer-bearing anchor for certain query types.

\paragraph{FlatTrans.}
\label{app:flattrans}
\textsc{FlatTrans} starts from the same factual anchors as \method{}.
It then collects all transition records associated with the user's active topic groups---both \intraedges{} and \interedges{}---and ranks them as flat documents by cosine similarity between the query embedding and each transition record embedding.
The top-$K_T$ transition records are merged with the factual anchors under the same final context budget and QA prompt as \method{}.
Unlike \method{}, \textsc{FlatTrans} does not follow transition endpoints, does not maintain a trajectory state, and does not adaptively expand across hops.
This tests whether transition text is sufficient once factual anchors are fixed, independently of graph-grounded endpoint traversal.

For example, if the query asks for a current cultural recommendation, \textsc{FlatTrans} may retrieve transition records mentioning cultural participation, but it does not require those records to be reachable from the factual anchors or to form a coherent endpoint-grounded update path.

On Qwen3-8B, \textsc{FlatTrans} reaches 54.32\% (32k) and 46.92\% (128k), which are the values reported alongside \textsc{ChronoExpand} and \method{} in Section~\ref{sec:hops}.

\subsection{Adaptive Retrieval: Per-Category Gain and Hop Statistics}
\label{app:adaptive-hops}

This section reports a finer-grained breakdown of where adaptive trajectory expansion helps.
For both Qwen3-8B and Qwen3.5-27B in the 128k setting, we compare the \emph{Facts Only} condition (factual anchor retrieval only, no expansion) with the full \method{} model and report the average adaptive hop depth per query category.

\paragraph{Qwen3-8B (128k).}
Table~\ref{tab:adaptive-hops-8b} provides the values underlying the main-text comparison in Figure~\ref{fig:per-category-8b}.
Overall accuracy increases from 46.00\% to 48.03\%, with an average expansion depth of 1.92 hops.

\begin{table}[h]
\centering
\caption{Per-category \emph{Facts Only} vs.\ \emph{Full} \method{} with average adaptive hop depth (128k, Qwen3-8B).}
\label{tab:adaptive-hops-8b}
\footnotesize
\setlength{\tabcolsep}{5pt}
\begin{tabular}{l r r r r}
\toprule
Category & Facts Only & Full & $\Delta$ & Avg Hops \\
\midrule
Revisit       & 46.84 & 51.90 & \textbf{+5.06} & 2.17 \\
Track Evo     & 51.82 & 55.00 & \textbf{+3.18} & 2.51 \\
Latest Prefs  & 53.12 & 56.75 & \textbf{+3.63} & 1.90 \\
Aligned Recs  & 51.29 & 51.29 & \phantom{+}0.00 & 1.72 \\
New Scenario  & 47.37 & 53.22 & \textbf{+5.85} & 1.81 \\
Shared Facts  & 57.79 & 62.99 & \textbf{+5.20} & 1.89 \\
New Ideas     & 24.40 & 21.13 & $-3.27$ & 1.77 \\
\midrule
\textit{Overall} & \textit{46.00} & \textit{48.03} & \textit{+2.03} & \textit{1.92} \\
\bottomrule
\end{tabular}
\end{table}

\begin{figure}[h]
\centering
\includegraphics[width=0.92\linewidth]{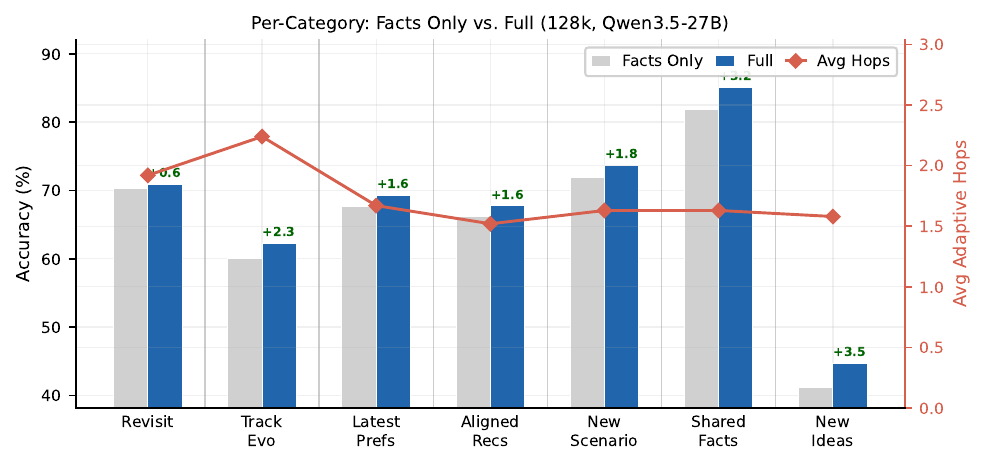}
\caption{Per-category comparison of \textit{Facts Only} versus \textit{Full} \method{} in the 128k setting (Qwen3.5-27B).
Bars show accuracy; the right axis shows the average adaptive hop depth of the full model.}
\label{fig:per-category-27b}
\end{figure}

\paragraph{Qwen3.5-27B (128k).}
On the stronger backbone, adaptive expansion yields positive deltas across all seven categories (Figure~\ref{fig:per-category-27b} and Table~\ref{tab:adaptive-hops-27b}).
Gains are largest for \emph{New Ideas} ($+3.48$\,pp), \emph{Shared Facts} ($+3.24$\,pp), and \emph{Track Evolution} ($+2.27$\,pp).
As at 8B, \emph{Track Evolution} and \emph{Revisit} trigger the deepest average expansions (2.24 and 1.92 hops), although the largest accuracy gain occurs on \emph{New Ideas}.
The 27B per-category pattern is broadly consistent with the 8B pattern for trajectory-sensitive categories, and the regression on \emph{New Ideas} observed at 8B does not appear at 27B, suggesting that the stronger backbone handles open-ended trajectory context more selectively.

\begin{table}[h]
\centering
\caption{Per-category \emph{Facts Only} vs.\ \emph{Full} \method{} with average adaptive hop depth (128k, Qwen3.5-27B).}
\label{tab:adaptive-hops-27b}
\footnotesize
\setlength{\tabcolsep}{5pt}
\begin{tabular}{l r r r r}
\toprule
Category & Facts Only & Full & $\Delta$ & Avg Hops \\
\midrule
Revisit       & 70.25 & 70.89 & \textbf{+0.64} & 1.92 \\
Track Evo     & 60.00 & 62.27 & \textbf{+2.27} & 2.24 \\
Latest Prefs  & 67.63 & 69.23 & \textbf{+1.60} & 1.67 \\
Aligned Recs  & 66.13 & 67.74 & \textbf{+1.61} & 1.52 \\
New Scenario  & 71.93 & 73.68 & \textbf{+1.75} & 1.63 \\
Shared Facts  & 81.82 & 85.06 & \textbf{+3.24} & 1.63 \\
New Ideas     & 41.18 & 44.66 & \textbf{+3.48} & 1.58 \\
\midrule
\textit{Overall} & \textit{62.56} & \textit{64.69} & \textit{+2.13} & \textit{1.74} \\
\bottomrule
\end{tabular}
\end{table}

\subsection{Construction-Stage Hyperparameter Selection}
\label{app:build-hypara}

This section reports three empirical studies on the construction stage: a validation of the topic router and LLM shift detector, a sensitivity analysis of the primary construction hyperparameters, and a quality analysis of the stored transition records.
All studies use the same 20-user diagnostic subset (257 queries) from PersonaMem 32k.
We first fix the construction parameters and then select the retrieval parameters on the same diagnostic subset.

\paragraph{Topic router and LLM shift detector.}

\method{} maintains a per-user memory graph organized into fine-grained topic subgraphs, isolating preference facts from semantically distinct life areas.
The topic router provides the provisional context for candidate shifts, while the shift detector compares buffered activity against a compact chronological evidence set that may include relevant facts outside the current topic.
To evaluate the contribution of each component, we compare four configurations that add topic structure and transition edges incrementally.

\begin{table}[h]
\centering
\caption{Validation of topic routing and transition edges (20-user diagnostic subset, 257 queries). Overall accuracy (\%).}
\label{tab:router-ablation}
\footnotesize
\setlength{\tabcolsep}{5pt}
\renewcommand{\arraystretch}{1.05}
\begin{tabular}{l c c r}
\toprule
Configuration & Topic Router & Trans.\ Edges & Acc (\%) \\
\midrule
Single topic, no transitions, no router & \ding{55} & \ding{55} & 51.75 \\
Single topic + transition edges (intra only)        & \ding{55} & \ding{51} & 54.86 \\
Multi-topic router, no transitions      & \ding{51} & \ding{55} & 55.64 \\
Full \method{} (multi-topic + transitions) & \ding{51} & \ding{51} & \textbf{57.59} \\
\bottomrule
\end{tabular}
\end{table}

The results in Table~\ref{tab:router-ablation} indicate that each component provides a positive incremental gain in this study and the combined system performs best.
Adding transition edges to a single-topic graph provides $+3.11$\,pp, suggesting that the LLM detector can identify useful preference-update signals even without per-topic isolation.
Adding the topic router without transition edges provides $+3.89$\,pp over the flat baseline, consistent with per-topic isolation reducing inter-topic interference in factual retrieval.
The full system, combining both components, achieves the highest accuracy ($+5.84$\,pp over the flat baseline).

\paragraph{Construction hyperparameter sensitivity.}
The construction stage is governed by three hyperparameters: the macro routing threshold $\tau_{\mathrm{macro}}$, the micro entity-consistency threshold $\tau_{\mathrm{micro}}$, and the shift-detection buffer size $\kappa$.
Due to the substantial offline cost of rebuilding topic graphs, this study uses the fixed 20-user diagnostic subset.

Table~\ref{tab:build-hypara} reports the key configurations.
A higher $\tau_{\mathrm{macro}}$ raises the bar for assigning a message to an existing topic, increasing \dbirth{} (topic creation) events and producing a more fragmented topic graph; this risks isolating semantically related facts into separate topics.
A lower $\tau_{\mathrm{micro}}$ suppresses evolution candidates and reduces the number of transition edges, which in turn limits the recall of preference-update signals available to the retriever.
With both routing thresholds fixed, buffer sizes $\kappa{=}5$ and $\kappa{=}20$ yield the same accuracy.
We adopt $\tau_{\mathrm{macro}}{=}0.65$, $\tau_{\mathrm{micro}}{=}0.25$, and $\kappa{=}5$.

\begin{table}[h]
\centering
\caption{Construction hyperparameter sensitivity (20-user diagnostic subset).}
\label{tab:build-hypara}
\footnotesize
\setlength{\tabcolsep}{5pt}
\renewcommand{\arraystretch}{1.05}
\begin{tabular}{c c c r}
\toprule
$\tau_{\mathrm{macro}}$ & $\tau_{\mathrm{micro}}$ & $\kappa$ & Acc (\%) \\
\midrule
\textbf{0.65} & \textbf{0.25} & \textbf{5}  & \textbf{56.81} \\
0.70          & 0.25          & 5            & 54.47 \\
0.65          & 0.20          & 10           & 56.42 \\
0.65          & 0.25          & 20           & 56.81 \\
\bottomrule
\end{tabular}
\end{table}

\paragraph{Transition-Record Quality Analysis.}
To assess the semantic quality of transition records produced by the shift detector, we sample stored records from the frozen PersonaMem 32k memory graphs and evaluate them with GPT-4o as an external judge.
This analysis focuses on whether stored records are meaningful and usable for retrieval, rather than estimating exhaustive recall over all possible preference updates in the raw history.
The judge prompt is shown in Appendix~\ref{app:prompts}.

\emph{Setup.}
We randomly sample 100 intra-topic and 100 inter-topic transition edges (200 total) with a fixed random seed and present each edge to GPT-4o as an independent judge.
An edge is rated \textsc{Valid} only if (1) old and new preference states are meaningfully different, (2) the trigger plausibly explains the change, and (3) the new state is specific enough to support future personalized retrieval.
An edge is \textsc{Invalid} if the change is trivial rewording, cessation-only without a replacement preference, based on an implausible trigger, incoherent, or too vague.

\emph{Results.}
Table~\ref{tab:shift-quality} reports the outcome.

\begin{table}[h]
\centering
\caption{LLM-judged quality of stored transition records.}
\label{tab:shift-quality}
\footnotesize
\setlength{\tabcolsep}{6pt}
\renewcommand{\arraystretch}{1.05}
\begin{tabular}{l r r}
\toprule
Record type & Sampled & Valid rate \\
\midrule
Intra-topic & 100 & 98.0\% \\
Inter-topic  & 100 & 99.0\% \\
\midrule
Total        & 200 & 98.5\% \\
\bottomrule
\end{tabular}
\end{table}

The high valid rate indicates that most stored transition records encode a meaningful old-to-new preference update with a plausible update rationale.
The three \textsc{Invalid} cases share the same failure mode: the detector promotes a cessation event into a transition record without a replacement preference, so the target entity does not encode a useful new preference state.

\emph{Positive example.}
User \texttt{aa95cf}, intra-topic (\texttt{craft supplies} $\to$ \texttt{crafting event}, 2015-03-23):
\emph{``Crafting Preference Shift: Disliking crafting and finding it tedious $\to$ Appreciating crafting as a therapeutic and creative activity (Trigger: Joining a DIY crafting group and creating personalized gifts).''}
Verdict: \textsc{Valid}. The before/after states are clearly distinct, the trigger is concrete, and the new state is specific enough to guide future recommendations.

\emph{Negative example.}
User \texttt{373af9}, intra-topic (\texttt{first aid techniques} $\to$ \texttt{anxiety}, 2015-01-21):
\emph{``Health Knowledge Shift: Interested in learning first aid $\to$ Discontinued pursuit of medical knowledge due to stress and anxiety.''}
Verdict: \textsc{Invalid}. The edge describes cessation without encoding a new preference direction; the target entity \texttt{anxiety} is not actionable for personalized retrieval.

\subsection{Inference-Stage Hyperparameter Selection}
\label{app:infer-hypara}

We study the main inference-stage hyperparameters of \method{} on a randomly selected sample of 20 users from PersonaMem 32k, comprising 257 evaluation queries.
To keep the search tractable, the retrieval configuration is tuned sequentially: we first vary the factual-anchor budget (Equation~\ref{eq:anchor}), then the trajectory-retention factor $\gamma$ (Equation~\ref{eq:state-update}), and finally the continuation/stopping rule $(\tau_c,\delta_c)$ (Equation~\ref{eq:path-continue}).
The resulting configuration is used for all main full-evaluation experiments.
For the 128k setting, the trajectory-related configuration chosen from this study is carried over, and only the factual-anchor budget is retuned to account for longer histories.

\begin{figure}[h]
\centering
\includegraphics[width=\linewidth]{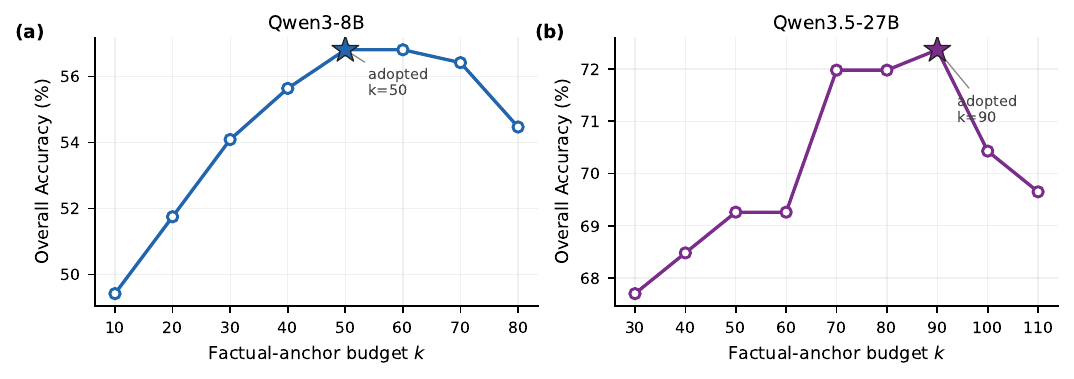}
\caption{Sensitivity to the factual-anchor budget $k$.
\textbf{(a)} Qwen3-8B; \textbf{(b)} Qwen3.5-27B.
The adopted value is marked with a star.}
\label{fig:infer-topk}
\end{figure}

\paragraph{Factual-anchor budget (top-$k$).}
Figure~\ref{fig:infer-topk} shows the accuracy curves for varying $k$ (Equation~\ref{eq:anchor}), with separate panels for Qwen3-8B ($k\in\{10,\dots,80\}$) and Qwen3.5-27B ($k\in\{30,\dots,110\}$) since the optimal range differs by backbone.
For Qwen3-8B, accuracy improves up to $k{=}50$ (56.81\%), with $k{=}60$ matching this peak before declining; we adopt $k{=}50$ as the smaller budget that matches the peak.
For Qwen3.5-27B, the curve peaks at $k{=}90$ (72.37\%), and we adopt this value directly.

\paragraph{Trajectory-retention factor ($\gamma$).}

\begin{figure}[h]
\centering
\includegraphics[width=0.86\linewidth]{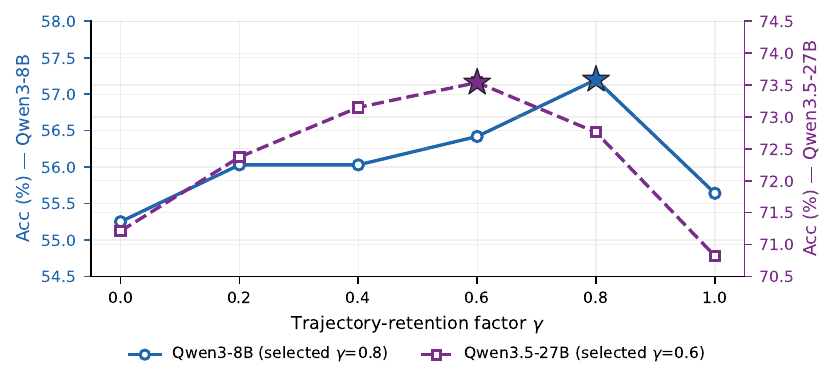}
\caption{Sensitivity to the trajectory-retention factor $\gamma$.
Two $y$-axes are used because the two backbones differ by roughly $17$\,pp on this subset.
The adopted value for each backbone is marked with a star.}
\label{fig:infer-gamma}
\end{figure}

The trajectory update in Equation~\ref{eq:state-update} can be unrolled as
\[
C_t
=
\gamma^t E(q)
+
(1-\gamma)\sum_{j=1}^{t}\gamma^{t-j}E(e_j).
\]
Thus, the score of a next-hop candidate is evaluated against a mixture of the original query and the accepted transition path, with exponentially decaying weight on earlier accepted transitions.
At $\gamma=1$, the trajectory state remains equal to the query embedding, so transition scoring is query-conditioned while graph traversal remains path-dependent.
When $\gamma=0$, the state is overwritten by the most recent transition, making expansion more vulnerable to local drift.
Intermediate values preserve query relevance while allowing accepted transition records to steer later hops, which is consistent with the sensitivity results in Figure~\ref{fig:infer-gamma}.
The results show accuracy as $\gamma$ varies across $\{0.0, 0.2, 0.4, 0.6, 0.8, 1.0\}$.
At $\gamma{=}0$, the trajectory state is overwritten by the most recently accepted edge, discarding all accumulated path information.
Intermediate values outperform both extremes: the 8B curve peaks at $\gamma{=}0.80$ (57.20\%) and the 27B curve at $\gamma{=}0.60$ (73.54\%).
This is consistent with the benefit of balancing accumulated preference context against newly accepted transition records at each hop.
We adopt $\gamma{=}0.80$ for Qwen3-8B and $\gamma{=}0.60$ for Qwen3.5-27B.

\paragraph{Continuation and drop tolerance ($\tau_c$, $\delta_c$).}
We evaluate four configurations spanning two levels of continuation threshold $\tau_c$ and drop tolerance $\delta_c$ (Equation~\ref{eq:path-continue}).
A lower $\tau_c$ is easier for a candidate edge to satisfy, allowing the retriever to continue to additional hops.
A lower $\delta_c$ enforces stricter tolerance against abrupt score drops, guarding against trajectory drift.
Table~\ref{tab:infer-stopgrid} reports the results.
The configuration $\tau_c{=}0.40$, $\delta_c{=}0.10$ performs best across both backbones: the permissive continuation threshold encourages multi-hop traversal while the strict drop tolerance prevents the trajectory from drifting toward unrelated edges.

\begin{table}[h]
\centering
\caption{Continuation/stopping sensitivity (Equation~\ref{eq:path-continue}).
Values are overall accuracy (\%). Best configuration in bold.}
\label{tab:infer-stopgrid}
\footnotesize
\setlength{\tabcolsep}{5pt}
\renewcommand{\arraystretch}{1.05}
\begin{tabular}{l c r}
\toprule
Backbone & $(\tau_c,\;\delta_c)$ & Acc\,(\%) \\
\midrule
\multirow{4}{*}{Qwen3-8B}
  & $\mathbf{(0.40,\;0.10)}$ & \textbf{57.59} \\
  & $(0.50,\;0.10)$ & 56.81 \\
  & $(0.50,\;0.20)$ & 54.86 \\
  & $(0.40,\;0.20)$ & 52.14 \\
\midrule
\multirow{4}{*}{Qwen3.5-27B}
  & $\mathbf{(0.40,\;0.10)}$ & \textbf{72.37} \\
  & $(0.50,\;0.20)$ & 71.98 \\
  & $(0.50,\;0.10)$ & 71.60 \\
  & $(0.40,\;0.20)$ & 69.26 \\
\bottomrule
\end{tabular}
\end{table}

\paragraph{Beam size and hop cap.}
Adaptive continuation in Algorithm~\ref{alg:infer} is controlled by $(\tau_c, \delta_c)$, while beam size $B{=}3$ and hop cap $H{=}6$ bound search width and depth.

\paragraph{Final retrieval configuration and 128k transfer.}
The adopted 32k configuration is $k{=}50$, $\gamma{=}0.80$, $\tau_c{=}0.40$, $\delta_c{=}0.10$ for Qwen3-8B, and $k{=}90$, $\gamma{=}0.60$, $\tau_c{=}0.40$, $\delta_c{=}0.10$ for Qwen3.5-27B.
For the 128k setting, the trajectory configuration is carried over and only the factual-anchor budget is retuned, yielding $k{=}90$ for Qwen3-8B and $k{=}110$ for Qwen3.5-27B.

\subsection{Case Analysis}
\label{app:case-analysis}

The categories where \method{} is not consistently strongest tend to involve either locally answerable latest-state lookup or open-ended ideation. In the former case, a strong note- or profile-based memory baseline can succeed by directly retrieving the most recent surface fact, while transition expansion may add little if the question does not require resolving a supersession relation. In the latter case, performance depends not only on memory recovery but also on the backbone's ability to creatively extrapolate from retrieved evidence. These cases suggest that transition records are most beneficial when the answer depends on determining which prior state has been replaced, rather than when the task is direct lookup or unconstrained generation.

We illustrate when adaptive transition expansion helps and when it can introduce noise through four PersonaMem cases.

\begin{tcolorbox}[
  colback=gray!3, colframe=black!35,
  title={\footnotesize\bfseries Case 1: Aligned Recommendation (\texttt{856b})},
  fonttitle=\footnotesize\bfseries, coltitle=black,
  boxrule=0.4pt, arc=1mm,
  left=4pt, right=4pt, top=3pt, bottom=3pt, breakable
]
\footnotesize
\textbf{Query.} Recommend a local weekend event that is both cultural and interactive.

\textbf{Options.} (a) interactive boutique art festival; (b) refusal; (c) literary festival; (d) food/drink tasting.

\textbf{Outcome.} Gold \texttt{(a)}; factual-anchor-only \texttt{(c)}; \method{} \texttt{(a)}.

\textbf{Representative factual anchors.}
\begin{itemize}
  \item Conducted an insightful interview with a local author.
  \item Is involved in organizing community events.
  \item Personalized recommendations aim to create engaging interaction.
\end{itemize}

\textbf{Accepted transition records.}
\begin{itemize}
  \item \emph{Passive literary engagement} $\rightarrow$ \emph{active literary-community participation} (intra-topic; rationale: national literature festival).
  \item \emph{Reading for personal enjoyment} $\rightarrow$ \emph{engaging with literature through interviews and creative-process exploration} (intra-topic; rationale: interview with a local author).
  \item \emph{Passive literature consumption} $\rightarrow$ \emph{active participation in literary communities and discussions} (inter-topic; rationale: book-signing events at indie bookstores).
\end{itemize}
\end{tcolorbox}

\noindent\textit{Interpretation.}
The factual anchors make the literary option plausible but remain isolated: they do not specify whether the user's current preference is passive literary attendance or participatory cultural interaction.
The accepted transition records shift the evidence toward active community-facing cultural participation, which is better aligned with the interaction-heavy art festival.
These records provide complementary evidence for the preference update.

\begin{tcolorbox}[
  colback=gray!3, colframe=black!35,
  title={\footnotesize\bfseries Case 2: Preference-Arc Recovery (\texttt{7d41})},
  fonttitle=\footnotesize\bfseries, coltitle=black,
  boxrule=0.4pt, arc=1mm,
  left=4pt, right=4pt, top=3pt, bottom=3pt, breakable
]
\footnotesize
\textbf{Query.} Inspired by a travel memoir, the user restarted emotionally rich travel journaling. Which option best describes the attitude toward travel?

\textbf{Options.} (a) \emph{disliking}$\to$\emph{liking}$\to$\emph{liking};
(b) \emph{loving}$\to$\emph{disliking}$\to$\emph{liking};
(c) \emph{no prior interest}$\to$\emph{liking}$\to$\emph{liking};
(d) \emph{liking}$\to$\emph{disliking}$\to$\emph{liking}.
\emph{Note: the original answer choices are long; we show the preference-trajectory keywords.}

\textbf{Outcome.} Gold \texttt{(d)}; factual-anchor-only \texttt{(b)}; \method{} \texttt{(d)}.

\textbf{Representative factual anchors.}
\begin{itemize}
  \item Started a travel journal to capture emotions and stories from adventures.
  \item Experiences anxiety when thinking about organizing a trip.
  \item Travel no longer brings the same excitement.
\end{itemize}

\textbf{Accepted transition record.}
\begin{itemize}
  \item \emph{Enjoying travel planning and exploring new places} $\rightarrow$ \emph{experiencing anxiety and difficulty with travel planning} (intra-topic; rationale: rising pressure from cooking and surprise-trip planning).
\end{itemize}
\end{tcolorbox}

\noindent\textit{Interpretation.}
The anchors expose current journal interest and earlier anxiety, while the transition explicitly links earlier enjoyment of travel planning to later anxiety.
Together, they support the \emph{liking$\to$disliking$\to$liking} arc in \texttt{(d)}.
The factual-anchor-only model instead selects \texttt{(b)}, which overstates the initial preference as \emph{loving} rather than \emph{liking}.

\begin{tcolorbox}[
  colback=gray!3, colframe=black!35,
  title={\footnotesize\bfseries Case 3: Expansion Noise on Fact Recall (\texttt{da4a})},
  fonttitle=\footnotesize\bfseries, coltitle=black,
  boxrule=0.4pt, arc=1mm,
  left=4pt, right=4pt, top=3pt, bottom=3pt, breakable
]
\footnotesize
\textbf{Query.} What are some fun ways to bring family together during the weekends?

\textbf{Options.} (a) family movie night; (b) garden project; (c) cooking challenge; (d) board-game tournament.

\textbf{Outcome.} Gold \texttt{(d)}; factual-anchor-only \texttt{(d)}; adaptive \method{} (8B) \texttt{(a)}.

\textbf{Representative factual anchor.}
\begin{itemize}
  \item Games are chosen as one of the activities to bring the family together.
\end{itemize}

\textbf{Accepted transition record (most salient).}
\begin{itemize}
  \item \emph{Awkward campfire gatherings} $\rightarrow$ \emph{family movie nights} (intra-topic; rationale: movie nights felt more comfortable and enriching).
\end{itemize}
\end{tcolorbox}

\noindent\textit{Interpretation.}
The factual anchors support the gold board-game option.
The accepted transition adds a competing ``family movie nights'' signal, and the 8B QA model selects that option after expansion.
Qwen3.5-27B answers correctly from the same context, suggesting that the backbone affects how the retrieved evidence is used.

\begin{tcolorbox}[
  colback=gray!3, colframe=black!35,
  title={\footnotesize\bfseries Case 4: Over-Grounding on Open-Ended Idea Generation (\texttt{0787})},
  fonttitle=\footnotesize\bfseries, coltitle=black,
  boxrule=0.4pt, arc=1mm,
  left=4pt, right=4pt, top=3pt, bottom=3pt, breakable
]
\footnotesize
\textbf{Query.} What new creative writing pursuits could I explore to further engage with other readers and reflect on literary works?

\textbf{Options.} (a) blog posts / literature podcast; (b) literary salons at home; (c) online writers' workshops; (d) poetry reading circle.

\textbf{Outcome.} Gold \texttt{(a)}; adaptive \method{} (8B) \texttt{(b)}.

\textbf{Representative factual anchors.}
\begin{itemize}
  \item Enjoys reading literary fiction.
  \item Values reflecting on books with others.
\end{itemize}

\textbf{Accepted transition records.}
\begin{itemize}
  \item \emph{Reading for personal enjoyment} $\rightarrow$ \emph{engaging with literature through interviews and creative-process exploration} (intra-topic; rationale: insightful interview with a local author).
  \item \emph{Passive consumption of literature} $\rightarrow$ \emph{active participation in literary communities and discussions} (inter-topic; rationale: book-signing events at indie bookstores).
\end{itemize}
\end{tcolorbox}

\noindent\textit{Interpretation.}
The transition records are coherent and locally faithful to the user's literary trajectory.
For an open-ended idea-generation query, however, they over-ground the model in familiar in-person literary-community formats, biasing the 8B backbone toward a conservative preference-consistent continuation (literary salons) rather than the more creative outward-facing formats in the gold answer.
This case supports the main-text observation that \emph{New Ideas} does not consistently benefit from deeper retrieval.
Partly off-topic retrieved context may also contribute to this error.

\end{document}